\pdfoutput=1

\documentclass[11pt]{article}

\usepackage[preprint]{acl}

\usepackage{times}
\usepackage{latexsym}
\usepackage{graphicx}
\usepackage{textcomp}
\usepackage{xcolor}
\usepackage{amsmath} 
\usepackage{booktabs} 
\usepackage{array}
\usepackage{graphicx} 
\usepackage{multirow}
\usepackage{color}
\usepackage{makecell} 
\usepackage{float}
\usepackage{subcaption}
\usepackage{colortbl}
\usepackage{placeins}
\usepackage{comment}
\usepackage{pifont}
\usepackage{url}

\usepackage[T1]{fontenc}

\usepackage[utf8]{inputenc}

\usepackage{microtype}

\usepackage{inconsolata}

\usepackage{graphicx}

\title{MM-Verify: Enhancing Multimodal Reasoning with Chain-of-Thought Verification}

\author{Linzhuang Sun$^{1}$\thanks{Equal Contribution.}, \ Hao Liang$^{2*}$, \ Jingxuan Wei$^{1}$, \ Bihui Yu$^{1}$, \ Tianpeng Li$^{3}$, \\
 {\bf Fan Yang$^{3}$\thanks{Corresponding Author}}, {\bf\ Zenan Zhou$^{3\dagger}$}, {\bf\ Wentao Zhang$^{2\dagger}$}   \\
  $^{1}$University of Chinese Academy of Sciences \\
  $^{2}$Peking University \\
  $^{3}$Baichuan Inc. \\
  \textit{sunlinzhuang21@mails.ucas.ac.cn, hao.liang@stu.pku.edu.cn} \\
  \textit{\{yangfan, zhouzenan\}@baichuan-inc.com, wentao.zhang@pku.edu.cn}
}

\begin{document}
\maketitle
\begin{abstract}
According to the Test-Time Scaling, the integration of External Slow-Thinking with the Verify mechanism has been demonstrated to enhance multi-round reasoning in large language models (LLMs). However, in the multimodal (MM) domain, there is still a lack of a strong MM-Verifier. In this paper, we introduce MM-Verifier and MM-Reasoner to enhance multimodal reasoning through longer inference and more robust verification. First, we propose a two-step MM verification data synthesis method, which combines a simulation-based tree search with verification and uses rejection sampling to generate high-quality Chain-of-Thought (COT) data. This data is then used to fine-tune the verification model, MM-Verifier. Additionally, we present a more efficient method for synthesizing MMCOT data, bridging the gap between text-based and multimodal reasoning. The synthesized data is used to fine-tune MM-Reasoner. Our MM-Verifier outperforms all larger models on the MathCheck, MathVista, and MathVerse benchmarks. Moreover, MM-Reasoner demonstrates strong effectiveness and scalability, with performance improving as data size increases. Finally, our approach achieves strong performance when combining MM-Reasoner and MM-Verifier, reaching an accuracy of 65.3 on MathVista, surpassing GPT-4o (63.8) with 12 rollouts. Our code is made available \url{https://github.com/Aurora-slz/MM-Verify}.
\end{abstract}

\section{Introduction}

\begin{figure}[t]
\centering
\includegraphics[width=0.48\textwidth]{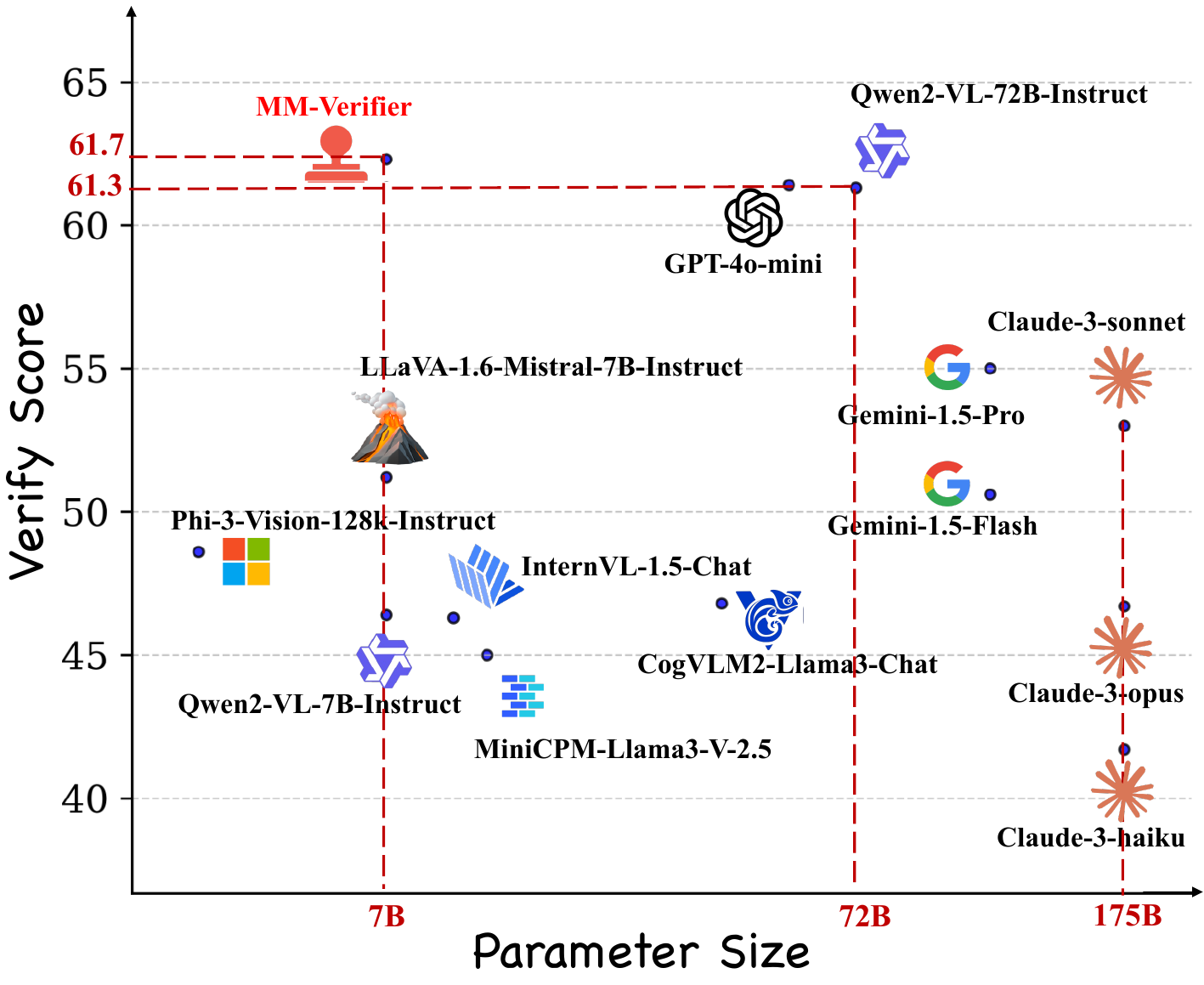}
\caption{Our 7B MM-Verifier outperform all other models, even large models like GPT-4o, Gemini and Claude on the MathCheck Outcome-Judging benchmark.}
\label{fig:fengmian}
\vspace{-4mm}
\end{figure}

Large language models (LLMs) have demonstrated exceptional performance across diverse tasks spanning myriad domains~\cite{chatgpt, llama}. Based on LLMs, MLLMs~\cite{zhao2023survey, wu2023multimodal, bai2024survey} also show strong understanding ability among different modalities~\cite{llava, bai2023qwenvl}. They have demonstrated strong performance in image classification~\cite{chen2024internvl}, image understanding~\cite{blip, li2023blip}, image captioning~\cite{bai2023qwenvl}, visual question answering~\cite{llava,llava1.5} and image-text retrieval~\cite{chen2024internvl}. Recently, MLLMs have also made significant strides in solving mathematical problems~\cite{liang2023unimath, huang2024hologram}. Researcher in our community have made efforts in designing strong reasoning models and algorithms~\cite{thawakar2025llamav, du2025virgo}. Despite the effort made in processing MLLMs, we still face two challenges:
\paragraph{Lack of Strong MM Verifiers} In pure-text LLMs, models can improve through self-critic methods~\cite{weng2022large, sun2024beats}. However, as shown in Table \ref{tab:mcts}, such methods face challenges in enhancing performance in multimodal models. Therefore, it is crucial to develop robust multimodal (MM) verifiers.

\paragraph{Lack of Long COT Reasoning Data} In the pure-text domain, models like DeepSeek-R1~\cite{guo2025deepseek}, s1~\cite{muennighoff2025s1}, and LIMO~\cite{ye2025limo} have demonstrated the effectiveness of Long COT data. However, in the multimodal domain, most collected mathematics problems are not in the Long COT format~\cite{li2024llava}. Therefore, the development of Long COT synthetic methods is necessary to enhance the reasoning ability of MLLMs.

To address these issues, in this paper, we introduce two novel data synthesis methods and subsequently train MM-Verifier and MM-Reasoner.  First, we perform a tree search, using simulations as rewards, to generate high-quality long MMCOT data. Next, we fine-tune multimodal large language models (MLLMs) on our data, resulting in long chain-of-thought (COT) responses. These data are then Given that MLLMs generate long COT responses, we then apply rejection sampling to further enhance their verification capabilities, leading to the proposal of MM-Verifier. After the introduction of MM-Verifier, we observed that long COT data significantly improves model reasoning performance. As a result, we aim to synthesize large amounts of long COT data to enhance the performance of base MLLMs. Since tree search can be computationally expensive, we use the MAVIS dataset~\cite{zhang2024mavis}, leveraging the descriptions of patterns in MAVIS and inputting them into the pure-text reasoning model, Qwen QWQ. We then pair these patterns with the corresponding responses from the Qwen QWQ model. This method has proven to be effective, scalable, and capable of efficiently constructing large amounts of long MMCOT data.

The core contributions are summarized as follows:
\begin{itemize}
    \item \textbf{MM Reasoning Data Synthesis Method} We propose two novel data synthesis methods for both our MM-Verifier and MM-Reasoner. First, we introduce a two-step MM-verification data synthesis approach that combines simulation-based tree search with GPT-4 verification and rejection sampling to generate high-quality COT data. Additionally, we use graphical software to link multimodal geometric shapes with textual descriptions, enabling the generation of multimodal reasoning data through a purely text-based reasoning model.
    
    \item \textbf{MM-Verifier} We introduce a new multimodal Outcome Reward Model (ORM) called MM-Verifier. MM-Verifier achieves state-of-the-art (SOTA) performance on the MathCheck benchmark, surpassing closed-source models such as GPT-4, Gemini, and Claude. Furthermore, our MM-Verifier-7B outperforms Qwen2-VL-72B across all metrics on the MathVista and MathVerse benchmarks.
    
    \item \textbf{Scalability of MM-Reasoner} We propose a novel MM-Reasoning model based exclusively on our synthetic COT data. Although our MM-Reasoner does not achieve SOTA performance, it outperforms the baseline models and demonstrates scalability as the size of the training dataset increases. This provides new insights for the development of more powerful MM-Reasoners.
    
    \item \textbf{Strong Performance} By combining MM-Verifier and MM-Reasoning, with a model size of only 7B parameters, we outperform both GPT-4 and human performance on the MathVista benchmark, highlighting the strong performance of our method.
\end{itemize}

\section{Related Work}

\subsection{MLLMs for Mathematics}
\paragraph{Commonly Used MLLMs}
The integration of visual knowledge into LLMs has become a pivotal area of research due to the rapid advancements in LLMs. MLLMs combine vision information from vision encoders with LLMs, thus enabling these models to process and interpret visual inputs for various visual tasks ~\cite{dino,glipv2, grounded-pt} with enhanced accuracy and efficiency. Pioneering frameworks like CLIP ~\cite{clip} leverage contrastive learning on expansive image-caption datasets to align modalities, forming the groundwork for cross-modal comprehension. Various adapters ~\cite{llava, llava1.5,  blip, blip1, pformer, lyrics} are introduced to further integrate different modalities. For example, LLaVA ~\cite{llava, llava1.5} employs a straightforward MLP to inject the vision information into LLMs. Whereas more complex implementations like the Q-Former in BLIP ~\cite{blip1, blip} utilize cross-attention to enhance modality integration. 

Recent studies~\cite{image-text-data, sharegpt4v, llava, llava1.5, otter, zhang2024critic, zhuang2024math, luo2025ursa} aim to enhance MLLM performance by improving the quality of both pre-training and fine-tuning datasets. Models such as LLaVA~\cite{llava, llava1.5}, ShareGPT4V~\cite{sharegpt4v}, LLaVA-Next, LLaVA-OneVision~\cite{llava, llava1.5}, Qwen2-VL, and Qwen2.5-VL~\cite{bai2023qwenvl} have demonstrated significant advancements in understanding and executing complex instructions through instruction tuning. Leveraging large-scale training data, these models have also achieved strong performance in solving mathematical problems~\cite{lu2023mathvista}.

\paragraph{MLLMs Designed for Math Problems}
In real-world applications, vision inputs are commonly used to present mathematical problems for models to solve. As a result, it is crucial for Vision-Language Large Models (MLLMs) to demonstrate strong mathematical capabilities. Meidani et al.~\cite{meidani2023snip} pioneered the use of symbolic data to train a Vision-Language Model (VLM) with mathematical proficiency. Building on this work, UniMath~\cite{liang2023unimath} combined vision, table, and text encoders with LLMs, achieving state-of-the-art performance at the time. Additionally, Huang et al.~\cite{huang2024hologram} succeeded in solving algebraic problems that involved geometric diagrams.

Another noteworthy line of research involves using LLMs to tackle geometric problems. G-LLaVA~\cite{gao2023g} fine-tuned LLaVA~\cite{llava} with geometric data, reaching SOTA performance in geometry. Subsequently, MAViS~\cite{zhang2024mavis} and EAGLE~\cite{li2024eagle} achieved SOTA results by introducing math-specific encoders and amassing large amounts of mathematical data.

\subsection{LLM-as-a-Judge}
In the Reinforcement Learning from Human Feedback (RLHF) or MCTS-based inference, Reward Models (RMs) are employed to assess and score the quality of model outputs, thereby guiding the optimization or reasoning path of LLMs. Reward models can be categorized into Process Reward Models (PRMs) and Outcome Reward Models (ORMs).

\textbf{Outcome Reward Models.} ORMs evaluate only the final mathematical results without considering the solution process. For instance, Qwen2.5-Math-RM-72B~\cite{zhang2025lessons}, released by the Qwen team, assigns a single score to each mathematical response.

\textbf{Process Reward Models.} PRMs are more fine-grained, focusing on whether each step of the reasoning path is logical and correct, providing step-level feedback and guidance signals. For example, Math-Shepherd~\cite{wang2024math} is trained on an automatically constructed (rather than manually annotated) process supervision dataset, scoring each step of mathematical reasoning. MATHMinos-PRM~\cite{gao2024llm} introduces a novel two-stage training paradigm and incorporates step-wise natural language feedback labels. EurusPRM~\cite{cui2025process} utilize implicit PRM, where ORM is trained to evaluate response-level labels. Qwen2.5-Math-PRM~\cite{zhang2025lessons}, currently the SOTA PRM, proposes a consensus filtering mechanism combining Monte Carlo estimation and LLM-as-a-judge. Additionally, there are the Skywork-PRM series~\cite{skyworkopeno12024} and RLHFlow-PRM series~\cite{wei2024implementation} models. Moreover,~\citet{liu2024diving} proposed Multimodal PRM based on Monte Carlo rollouts. For more comprehensive LLM-as-a-Judge please refer to the LLM-as-a-Judge survey~\cite{gu2024survey}.

\section{Methodology}

\begin{figure*}[t]
\centering\textbf{}
\includegraphics[width=\textwidth]{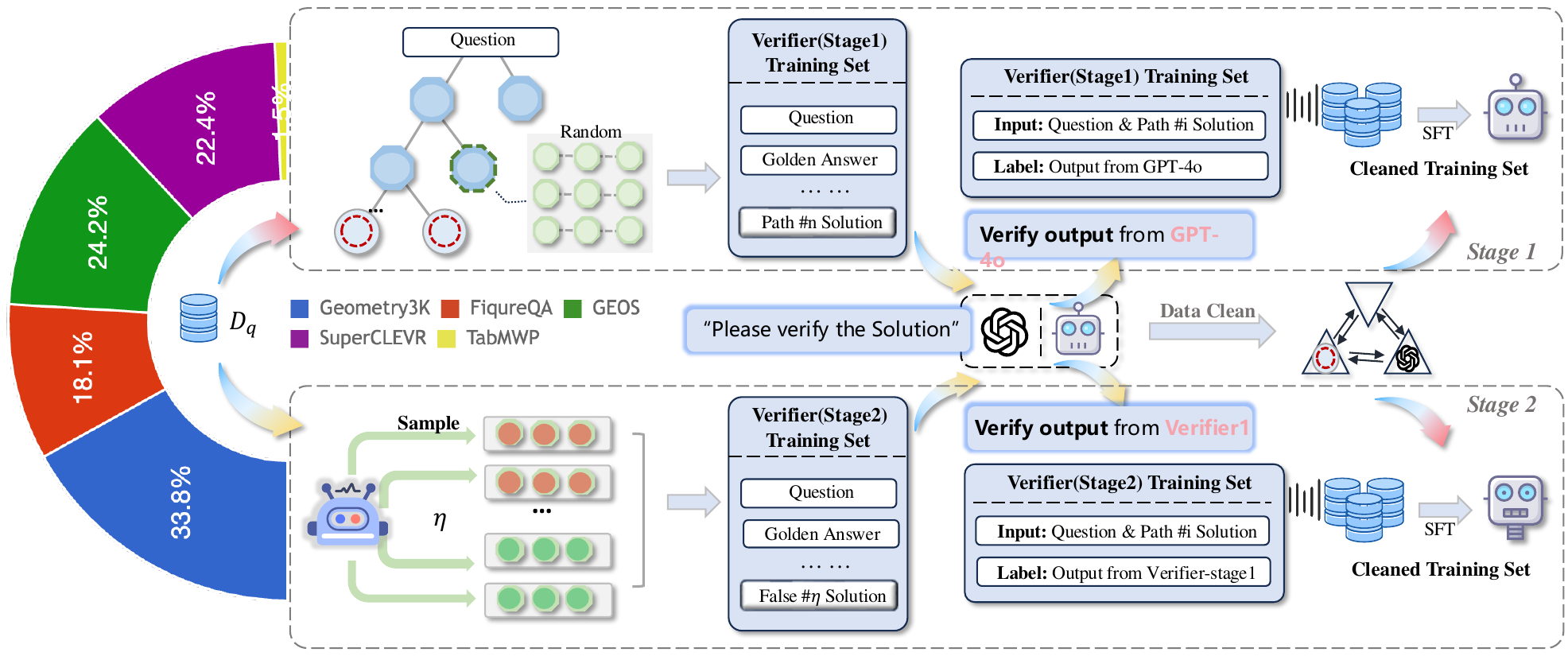}
\caption{We present the pipeline for synthesizing MM-Verifier data. In Stage 1, we use a simulation-based algorithm for long-chain CoT reasoning and long verification. In Stage 2, we use the trained Verifier model from Stage 1 to further enhance it using rejection sampling, generating more long CoT verification data.}
\label{fig:method}
\vspace{-4mm}
\end{figure*}

\begin{table}[htbp]
  \centering
  \caption{Statistical details of our collected data.}
  \resizebox{\linewidth}{!}{
    \begin{tabular}{llcc}
    \toprule
    \textbf{Dataset} & \textbf{Subdataset} & \textbf{Number} & \textbf{Ratio} \\
    \midrule
    \multirow{6}[4]{*}{MM-Verify} & Geometry3K & 20226 & 33.84\% \\
          & FigureQA & 10800 & 18.07\% \\
          & GEOS  & 882   & 1.48\% \\
          & Super-CLEVR & 14446 & 24.17\% \\
          & TabMWP & 13418 & 22.45\% \\
\cmidrule{2-4}          & sum   & 59772 & 100\% \\
    \midrule
    MM-Reasoner & MAVIS-Geo & 32146 & 100\% \\
    \bottomrule
    \end{tabular}%
    }
  \label{tab:dataset}%
  \vspace{-2mm}
\end{table}%

In this section, we introduce the construction process of MM-Verifier, as illustrated in Figure~\ref{fig:method}. Section~\ref{sec:3-1} details the data synthesis methodology for MM-Verifier (Stage 1). Section~\ref{sec:3-2} describes the data synthesis scheme employed in MM-Verifier (Stage 2). In Section~\ref{sec:3-3}, we explore the enhancement of multimodal model reasoning capabilities through the integration of long-COT data in pure text form.

\subsection{Stage1: Long CoT MM-Verifier}
\label{sec:3-1}
\subsubsection{Source Data Collection}  
MM-Verifier is designed to verify the correctness of a \( <q,s> \) pair by determining whether the solution \( s \) is correct. To achieve this goal, it is essential to synthesize diverse verification data from a variety of sources. Specifically, we construct the question source pool \( D_s \) from seven categories in MATH360V: Geometry3K, TabMWP, Super-CLEVR, UniGeo, FigureQA, and GEOS, with their statistical details summarized in Table~\ref{tab:dataset}. By collecting data from these categories, we obtained a diverse set of multimodal mathematics questions. 

However, solutions in these datasets are typically very short, and many contain only answers. When verifying reasoning, the answer is often provided in a long COT form. Therefore, we need to construct COT data ranging from short to long, along with their corresponding verifications. To facilitate the generation of long COT data and enhance the diversity of the training set, we design a simulation-based search algorithm to create extended reasoning trajectory data for training MM-Verifier.
\begin{figure}[t]
\centering
\includegraphics[width=0.47\textwidth]{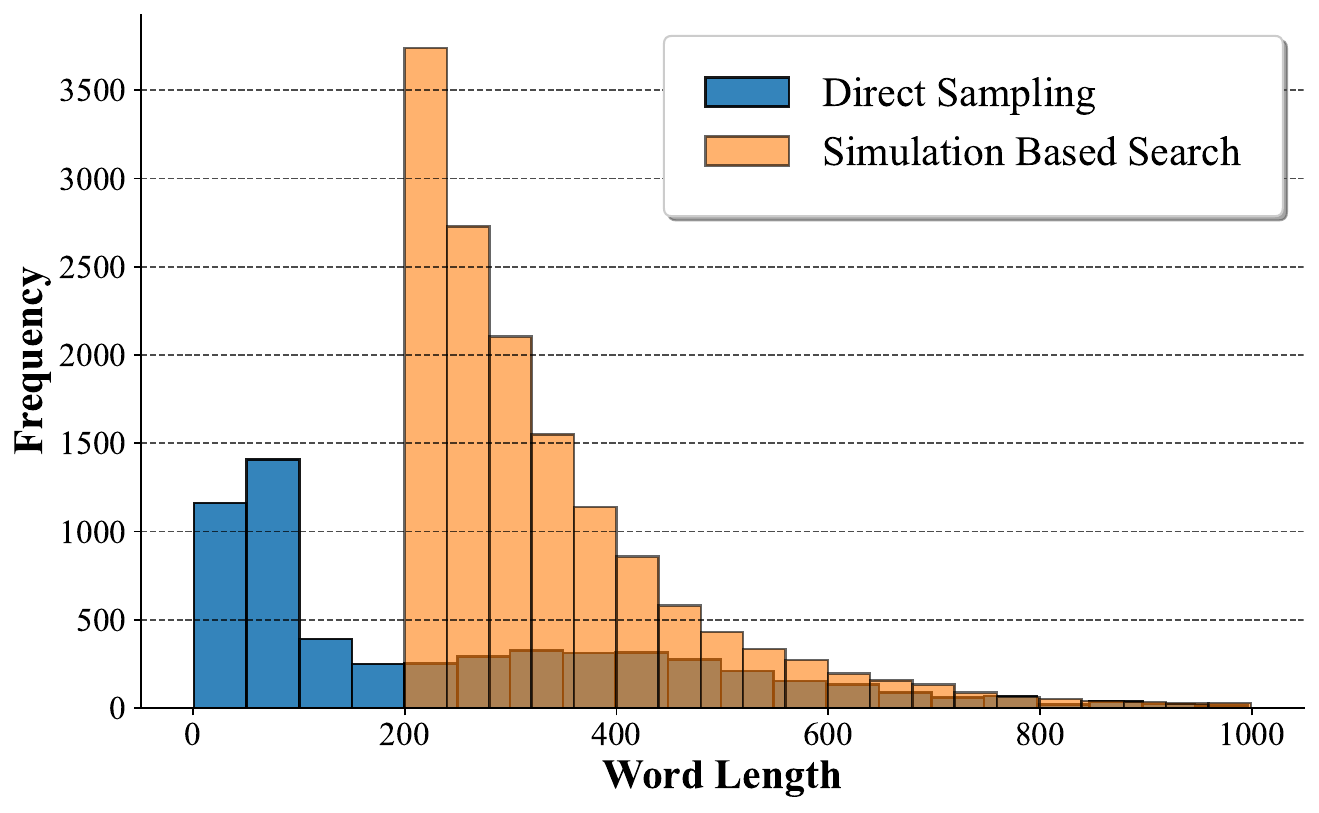}
\caption{Answer length of direct sampling and simulated-based search. We can see the simulated-based search can synthesize longer COT answers.}
\label{fig:lengthDist}
\vspace{-4mm}
\end{figure}
\subsubsection{Simulation-based Search Algorithm}
Inspired by Monte Carlo Tree Search (MCTS), we propose a simulation-based search algorithm tailored for multimodal models. However, we do not directly apply the traditional MCTS tree search, as multimodal models often fail to generate reliable rewards, which results in suboptimal performance, as illustrated in Appendix \ref{app:MCTS_performance}. To address this challenge, we introduce a simulation-based reward mechanism. 

Starting from the root node \( q_i \), we first simulate \( k \) child nodes for each node. For each child node, we perform simulations where the model directly generates answers based on the current node. For a tree node \( u_d \), which represents a node at depth \( d \), the ancestral path leading up to the root is denoted by the sequence \( \{ u_{d-1}, \dots, u_1 \} \). The simulation answer for this path is given by:
\[
\textit{Simulation Answer} = LLM\left(\bigoplus_{i=1}^{d-1} u_i\right)
\]
These simulations are repeated \( l \) times, and the correctness ratio of the \textit{Simulation Answer} is used as the reward. Once the reward is obtained, we apply the MCTS algorithm for further simulation and data synthesis.

Using the simulation-based MCTS approach, for each question, we perform \( n \) rollouts and collect solution pairs \( \langle q_i, p_j^i \rangle \), where \( j \in \{ 1, 2, \dots, n \} \) and \( n \) represents the number of leaf nodes. These \( n \) solution pairs are then verified as positive and negative cases for training the MM-verifier.


\begin{table*}[htbp]\small
  \centering
  \caption{We compare performance of Qwen2-VL-Instruct-7B/72B, LLaMA-3.2-11B-Vision-Instruct with our MM-Reasoner on the MathVista testmini benchmark. GVQA: General VQA; MVQA: Math Target VQA.}
  \resizebox{1.0\linewidth}{!}{
    \begin{tabular}{l||c|c|c||c|c|c||c|c|c}
    \toprule
    \multicolumn{1}{c||}{\multirow{2}[2]{*}{\textbf{Method}}} & \multicolumn{3}{c||}{\textbf{Qwen2-VL}} & \multicolumn{3}{c||}{\textbf{LLaMA-3.2-11B-Vision}} & \multicolumn{3}{c}{\textbf{MM-Reasoner}} \\
\cmidrule{2-10}          & \textbf{ALL} & \textbf{GVQA} & \textbf{MVQA} & \textbf{ALL} & \textbf{GVQA} & \textbf{MVQA} & \textbf{ALL} & \textbf{GVQA} & \textbf{MVQA} \\
    \midrule
    \rowcolor[rgb]{ .851,  .882,  .949} \multicolumn{10}{c}{\footnotesize \textit{Sample 4}} \\
    \midrule
    Majority Voting & 57.1  & 65.7  & 49.8  & 45.2  & 50.8  & 40.4  & 59.4  & 65.2  & 54.3 \\
    Qwen2-VL-7B as Judgement & 57.9  & 66.3  & 50.7  & 41.7  & 48.9  & 35.5  & 53.1  & 59.8  & 47.4 \\
    Qwen2-VL-72B as Judgement & 53.4 & 60.4 & 47.4 & 46.3 & 49.8 & 43.3 & 53.7 & 59.8 & 48.5 \\
    MM-Verifier(Stage1) & 58.8  & 64.1  & 54.3  & 50.0  & 56.3  & 44.6  & 60.6  & 66.5  & 53.7 \\
    MM-Verifier(Stage2) & \textbf{59.8} & \textbf{67.0} & \textbf{53.7} & \textbf{50.4} & \textbf{55.9} & \textbf{45.7} & \textbf{61.5} & \textbf{65.2} & \textbf{58.3} \\
    \midrule
    \rowcolor[rgb]{ .851,  .882,  .949} \multicolumn{10}{c}{\footnotesize \textit{Sample 8}} \\
    \midrule
    Majority Voting & 61.1  & \textbf{68.5}  & 54.8  & 48.3  & 52.8  & 44.4  & 62.2  & 68.0  & 57.2 \\
    Qwen2-VL-7B as Judgement & 54.5  & 62.0  & 48.1  & 46.1  & 50.7  & 42.2  & 53.6  & 61.5  & 46.9 \\
    Qwen2-VL-72B as Judgement & 56.2 & 62.4 & 50.9 & 46.2 & 51.1 & 42.0 & 53.9 & 61.1 & 47.8 \\
    MM-Verifier(Stage1) & 61.6  & 66.5  & 57.4  & 51.4  & 56.3  & 47.2  & 63.8  & 68.7  & 59.6 \\
    MM-Verifier(Stage2) & \textbf{62.5} & \textbf{68.5} & \textbf{57.4} & \textbf{52.1} & \textbf{57.6} & \textbf{47.4} & \textbf{65.3} & \textbf{69.8} & \textbf{61.5} \\
    \midrule
    \rowcolor[rgb]{ .851,  .882,  .949} \multicolumn{10}{c}{\footnotesize \textit{Sample 12}} \\
    \midrule
    Majority Voting & 62.9  & 68.5  & 58.1  & 51.3  & 56.5  & 46.9  & 64.8  & 68.7  & \textbf{61.5} \\
    Qwen2-VL-7B as Judgement & 54.4  & 60.0  & 49.6  & 45.5  & 48.6  & 42.0  & 55.4  & 61.1  & 50.6 \\
    Qwen2-VL-72B as Judgement & 55.7 & 61.1 & 51.1 & 46.5 & 49.8 & 43.7 & 55.6 & 61.5 & 48.7 \\
    MM-Verifier(Stage1) & 63.7  & 70.2  & 58.1  & 55.0  & 59.8  & 50.9  & 64.3  & 69.1  & 60.1 \\
    MM-Verifier(Stage2) & \textbf{64.1}  & \textbf{70.4}  & \textbf{58.7}  & \textbf{55.9}  & \textbf{60.9}  & \textbf{51.7}  & \textbf{65.2}  & \textbf{69.8}  & 61.3 \\
    \bottomrule
    \end{tabular}%
    }
  \label{tab:abla_mathvista}%
  \vspace{-4mm}
\end{table*}%

\subsubsection{Long COT Verification Data Synthesize}
After obtaining the \( \langle q_i, p_j^i \rangle \) pair, the next step is to verify \( p_j^i \) to determine whether \( q_i \) has been answered correctly. To do this, we use GPT-4o (gpt-4o-2024-08-06) to verify \( \langle q_i, p_j^i \rangle \) using the instruction "Verify step by step..." (for the detailed prompt, see Figure~\ref{fig:prompt-verify}). The model's output, denoted as \( v_i \), serves as the target for the verifier. The resulting dataset collected at this stage is represented as \( D_v \), which can be expressed as:
\[
D_v = \{(q_i, p_j^i, v_i) \mid i = 1, \dots, m; \, j = 1, \dots, n \}
\]
Additionally, we implement a data-cleaning strategy to filter high-quality synthetic data for training the MM-Verifier. For each data instance \( (q_i, p_j^i, v_i) \) in \( D_v \), we first design an answer extraction prompt, as shown in Figure \ref{fig:prompt-extract}) and use LLaMA-3.2-3B-Instruct for answer extraction, denoted by \texttt{extract()} for clarity. We then apply the following criteria for data cleaning:
\begin{itemize}
    \item \textbf{Condition 1:} If \( \text{extract}(p_j^i) \) matches the golden label \( \text{extract}(y_i) \), and the final result of \( v_i \) is the answer is correct.
    \item \textbf{Condition 2:} If \( \text{extract}(p_j^i) \) does not match \( \text{extract}(y_i) \), and the final result of \( v_i \) is the answer is not correct.
\end{itemize}
We collect the instances that meet the above conditions, \( (q_i, p_j^i, v_i) \), into \( D_{\text{clean}} \). Any instance that does not satisfy these conditions is discarded. The dataset \( D_{\text{clean}} \) is then used for supervised fine-tuning (SFT) on Qwen2-VL-7B-Instruct, yielding the first-stage verifier, MM-Verifier (Stage 1).


\subsection{Stage 2: Rejection Sampling Further Improves Verification}
\label{sec:3-2}

In Stage 1, we obtained a Verifier with strong long-chain Chain-of-Thought (CoT) reasoning capabilities. In Stage 2, our goal is to improve the efficiency of the data synthesis process, reduce API costs, and further enhance the Verifier's capabilities.

To achieve this, we first generate corresponding solutions based on a given set of questions. By leveraging the long CoT reasoning ability of the MM-Verifier from Stage 1, we can generate additional long CoT verification data.

The synthetic data is then cleaned using string matching against the correct answer. The filtered data is subsequently fed into the MM-Verifier (Stage 1) for further training, resulting in the enhanced MM-Verifier (Stage 2).
\begin{table*}[htbp]
  \centering
  \caption{We compare performance of Qwen2-VL-Instruct-7B/72B, LLaMA-3.2-11B-Vision-Instruct with our MM-Reasoner on the MathVerse testmini benchmark. VD: Vision Dominant; VI: Vision Intensive; TL: Text Lite.}
  \resizebox{1.0\linewidth}{!}{
    \begin{tabular}{l|c|c|c|c|c|c|c|c|c|c|c|c}
    \toprule
    \multicolumn{1}{c|}{\multirow{2}[4]{*}{\textbf{Method}}} & \multicolumn{4}{c|}{\textbf{Qwen2-VL}} & \multicolumn{4}{c|}{\textbf{llama-3.2-11B-Vision}} & \multicolumn{4}{c}{\textbf{MM Reasoner}} \\
\cmidrule{2-13}          & \textbf{ALL} & \textbf{VD} & \textbf{VI} & \textbf{TL} & \textbf{ALL} & \textbf{VD} & \textbf{VI} & \textbf{TL} & \textbf{ALL} & \textbf{VD} & \textbf{VI} & \textbf{TL} \\
    \midrule
    \rowcolor[rgb]{ .851,  .882,  .949} \multicolumn{13}{c}{ \textit{Sample  4}} \\
    \midrule
    Majority Voting & 20.1  & 19.8  & 19.0  & 17.7  & 17.8  & 15.9  & 17.4  & 19.8  & 22.9  & 20.7  & 23.0  & 23.2 \\
    Qwen2-VL-7B as Judgement & 20.6  & 22.3  & 20.8  & 19.0  & 20.6  & 19.3  & 21.1  & 21.2  & 22.6  & 21.4  & 21.4  & 24.6 \\
    Qwen2-VL-72B as Judgement  & 21.5 & 20.7 & 21.1 & 22.0 & 20.2 & 20.8 & 18.1 & 21.6 & 23.0& 22.0& 22.3& 24.0\\
    MM-Verifier(Stage1) & 24.0  & \textbf{25.3}  & 22.6  & 22.8  & 21.9  & \textbf{23.2}  & 20.8  & 22.5  & 24.8  & 22.2  & \textbf{25.6}  & 26.0 \\
    MM-Verifier(Stage2) & \textbf{24.3} & 23.6 & \textbf{23.2} & \textbf{23.1} & \textbf{22.4} & 21.8 & \textbf{23.0} & \textbf{24.4} & \textbf{25.3} & \textbf{23.1} & 23.5 & \textbf{27.4} \\
    \midrule
    \rowcolor[rgb]{ .851,  .882,  .949} \multicolumn{13}{c}{ \textit{Sample  8}} \\
    \midrule
    Majority Voting & 22.9  & 21.8  & 22.1  & 23.5  & 23.5  & 20.6  & 22.8  & 25.5  & 24.5  & 20.6  & 25.0  & 24.5 \\
    Qwen2-VL-7B as Judgement & 21.1  & 20.6  & 20.7  & 21.1  & 21.1  & 19.7  & 21.1  & 20.0  & 21.7  & 19.2  & 24.6  & 20.3 \\
    Qwen2-VL-72B as Judgement  & 21.5 & 20.3 & 21.2 & 21.2 & 20.8& 19.5 & 20.2 & 20.4& 22.3& 19.4& 24.4& 23.1\\
    MM-Verifier(Stage1) & 25.1  & 22.7  & \textbf{24.2}  & \textbf{25.5}  & 24.8  & 21.6  & 24.4  & 25.4  & 25.3  & 23.1  & 23.5  & \textbf{27.4} \\
    MM-Verifier(Stage2) & \textbf{25.2}  & \textbf{24.2}  & 24.1  & 25.1  & \textbf{25.0}  & \textbf{22.7}  & \textbf{26.3}  & \textbf{25.9}  & \textbf{25.7}  & \textbf{23.2}  & \textbf{25.3}  & 26.3 \\
    \midrule
    \rowcolor[rgb]{ .851,  .882,  .949} \multicolumn{13}{c}{ \textit{Sample  12}} \\
    \midrule
    Majority Voting & 24.8  & 19.9  & 24.7  & 25.6  & 24.0  & 19.7  & 24.5  & 26.1  & 25.9  & 23.1  & 26.1  & 27.3 \\
    Qwen2-VL-7B as Judgement & 22.0  & 20.1  & 22.0  & 22.5  & 20.0  & 18.9  & 19.9  & 23.1  & 21.8  & 19.5  & 22.7  & 23.4 \\
    Qwen2-VL-72B as Judgement  & 22.3 & 20.4 & 22.3 & 23.0 & 20.7 & 19.4 & 20.9 & 20.7 & 22.1 & 20.1 & 23.2 & 25.0\\
    MM-Verifier(Stage1) & 25.7  & \textbf{23.1}  & \textbf{26.5}  & 26.5  & 24.5  & 21.1  & 23.2  & 25.5  & 27.0  & 23.5  & 27.0  & \textbf{29.4} \\
    MM-Verifier(Stage2) & \textbf{25.8}  & 21.7  & 25.4  & \textbf{28.2}  & \textbf{24.7}  & \textbf{21.2}  & \textbf{24.7}  & \textbf{27.0}  & \textbf{27.3}  & \textbf{24.1}  & \textbf{27.7}  & 28.9 \\
    \bottomrule
    \end{tabular}%
    }
  \label{tab:abla_mathverse}%
  \vspace{-4mm}
\end{table*}%

\subsection{Bridging the Gap Between Text and MM}
\label{sec:3-3}
When applying MM-Verifier, the base model generates multiple reasoning paths for a given question. The MM-Verifier then evaluates these paths, distinguishing between correct and incorrect inferences. This process inherently requires the base model to produce at least one plausible correct reasoning path for the MM-Verifier to recognize and validate. Therefore, in addition to a promising MM-Verifier,In this section, our goal is to synthesize long COT data to train a robust MM-Reasoner capable of learning long COT reasoning. However, synthesizing long COT data using tree search can be computationally expensive. Moreover, long COT pure text models have demonstrated strong performance. To efficiently generate long COT data, we propose distilling knowledge from pure text models to our MM-Reasoner.

Specifically, we select data from MAVIS-GEOMETRY~\cite{zhang2024mavis}, which includes geometric pattern drawings along with textual instructions. By combining the geometric textual instructions with the original questions, we can feed them into a pure text reasoning model. The outputs generated by the reasoning model, Qwen-QwQ~\cite{qwq-32b-preview} with prompts in Figure \ref{fig:prompt-qwq}, are then collected as target labels for the MM-Reasoner training dataset, denoted as \( D_r \). To ensure the quality of the data, we filter out incorrect QwQ-generated reasoning results from \( D_r \). Then we use the filtered data to train Qwen2-VL-7B-Instruct with supervised fine-tuning (SFT), ultimately obtaining the MM-Reasoner model.

\section{Experiments}
\subsection{Experiment Setting}
\paragraph{Baseline Models.}  
The base models for MM-Verifier and MM-Reasoner are Qwen2-VL-Instruct-7B~\cite{bai2023qwenvl}. For comparison, we include random selection and human performance as baselines, along with two types of closed-source models: GPT-4o (gpt-4o-2024-08-06)~\cite{openai2023gpt} and Qwen-VL-Plus~\cite{bai2023qwen}. For open-source models, we evaluate ten MLLMs: mPLUG-Owl2-7B~\cite{ye2024mplug}, MiniGPT4-7B~\cite{zhu2023minigpt}, LLaVA-1.5-13B~\cite{llava1.5}, SPHINX-V2-13B~\cite{lin2023sphinx}, Deepseek-VL~\cite{lu2024deepseek}, LLaVA-OneVision-7B (llava-onevision-qwen2-7b-ov-hf)~\cite{li2024llava}, Qwen2-VL-Instruct-7B~\cite{bai2023qwenvl}, Llama-3.2-11B-Vision~\cite{llama}, Math-LLaVA~\cite{shi2024math}, and G-LLaVA-7B~\cite{gao2023g}.
\begin{figure*}[t]
\centering
\includegraphics[width=\textwidth]{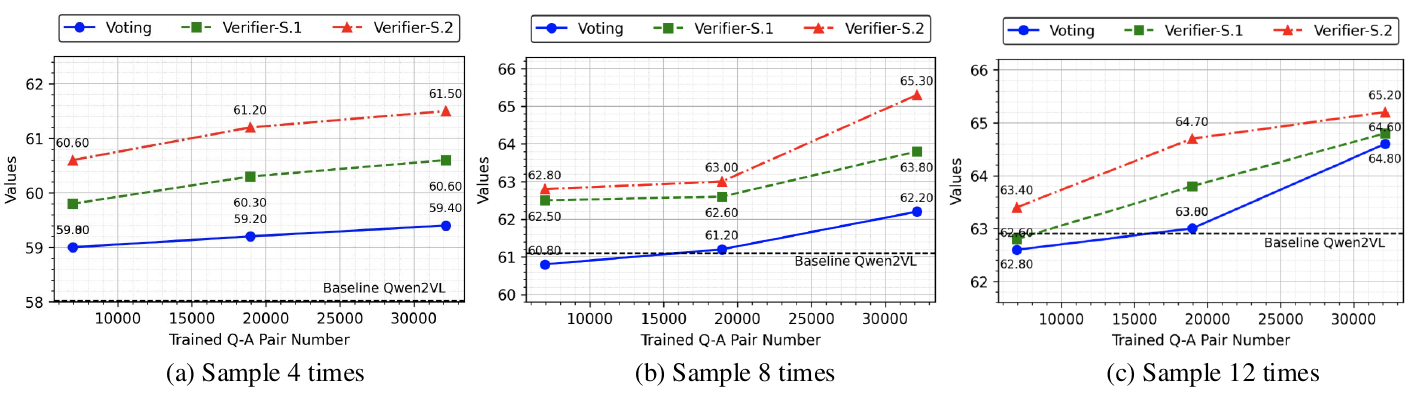}
\caption{The performance of our MM-Reasoner can scale up using different MM-Verifiers. We can see with different scale MM-Reasoner the MM-Verifier consistently outperform majority voting and MM-Verifier Stage1.}
\label{fig:mm_scale}
\vspace{-4mm}
\end{figure*}
\paragraph{MM-Verifier Baselines.}  
\textbf{Majority Voting:} Select the answer that appears most frequently among multiple candidate answers.  
\textbf{Qwen2-VL-7B and Qwen-VL-72B as Judgment:} We use Qwen2-VL-7B-Instruct and Qwen-VL-72B~\cite{bai2023qwenvl} as the judgment model to assess the correctness of each candidate solution. If multiple candidates are deemed correct, we apply a majority voting mechanism to determine the final answer.  

\paragraph{Benchmarks.}  
We evaluate the performance of MM-Verifier, MM-Reasoner, and the baselines on two commonly used benchmarks: MathVista~\cite{lu2023mathvista} and MathVerse~\cite{zhang2024mathverse}. Additionally, we assess the MM-Verifier on the Verify bench MathCheck (Multimodal Outcome Judging)~\cite{zhou2024your}.

\paragraph{Settings.}  
The settings include a maximum token limit of 4096, a top-k value of 5, a temperature of 0.3, and a repetition penalty of 1.05. All experiments are conducted on 8 NVIDIA H20 GPUs.

\subsection{Effectiveness of MM-Verifier and MM-Reasoner}
In this section, we demonstrate the effectiveness of both the MM-Verifier and MM-Reasoner. As shown in Figure \ref{fig:fengmian}, the MM-Verifier outperforms all other models in the MathCheck benchmark. Moreover, Tables \ref{tab:abla_mathvista} and \ref{tab:abla_mathverse} reveal that the MM-Verifier achieves strong performance on the MathVista and MathVerse benchmarks. Our results indicate that the MM-Verifier surpasses both the Majority Voting and Qwen2-VL-72B-Instruct methods. Specifically, MM-Verifier (Stage 2) delivers superior results, demonstrating its robust ability to verify answers and enhance model performance.

Furthermore, the MM-Reasoner outperforms powerful models, such as Qwen2-VL-Instruct-7B and LLaMA-3.2-11B-Vision, across all evaluation metrics. These findings clearly demonstrate that both the MM-Verifier and MM-Reasoner contribute significantly to performance improvements, underscoring their potential for addressing complex multimodal reasoning and verification tasks.

Interestingly, we observe that the performance of Qwen2-VL-72B, when used as a Judgment model, improves when verifying answers for Qwen2-VL and LLaMA-3.2-11B-Vision (from sample 4 to sample 12). However, its performance drops when verifying MM-Reasoner outputs. This discrepancy arises because conventional models struggle to verify the correctness of longer outputs, while our MM-Verifier consistently maintains robust performance. This further emphasizes the versatility of the MM-Verifier across a wide range of scenarios.

\subsection{Scalability of Our MM-Reasoner}

The results in Figure~\ref{fig:mm_scale} illustrate the scalability of MM-Reasoner with respect to the quantity of training data. As the amount of training data increases, the performance of the model consistently improves across all evaluation settings. Specifically, MM-Reasoner achieves steady performance gains when moving from 6,952 to 32,146 training samples, demonstrating its ability to effectively utilize larger datasets for better reasoning.

Additionally, both Verifier-S.1 and Verifier-S.2 show clear improvements as the training data grows, with Verifier-S.2 outperforming Verifier-S.1 in all cases. This trend highlights the effectiveness of the staged verification approach in enhancing reasoning accuracy.

These results emphasize the superiority of our method, as MM-Reasoner scales effectively with increasing training data, achieving higher performance and showcasing the robustness and adaptability of our approach.
\begin{table*}[htbp]\scriptsize
  \centering
  \caption{We compare our MM-Verifier plus MM-Reasoner with closed and open source MLLMs, and other baselines.  GPS: geometry problem solving; MWP: math word problem; FQA: figure question answering; TQA: textbook question answering; VQA: visual question answering.}
  \resizebox{1.0\linewidth}{!}{
    \begin{tabular}{l||c|c|c|c|c|c||c|c|c|c|c}
    \toprule
    \multicolumn{1}{c||}{\multirow{2}[4]{*}{\textbf{Model}}} & \multicolumn{6}{c||}{\textbf{MATHVISTA}}       & \multicolumn{5}{c}{\textbf{MATHVERSE}} \\
\cmidrule{2-12}          & \textbf{ALL} & \textbf{GPS} & \textbf{MWP} & \textbf{FQA} & \textbf{TQA} & \textbf{VQA} & \textbf{ALL} & \textbf{TD} & \textbf{TL} & \textbf{VI} & \textbf{VD} \\
    \midrule
    \rowcolor[rgb]{ .851,  .882,  .949} \multicolumn{12}{c}{\textit{Closed Source MLLMs \& Other Baselines}} \\
    \midrule
    Random & 17.9  & 21.6  & 3.8   & 18.2  & 19.6  & 26.3  & 12.4  & 12.4  & 12.4  & 12.4  & 12.4 \\
    Human & 60.3  & 48.4  & 73.0  & 59.7  & 63.2  & 55.9  & 64.9  & 71.2  & 70.9  & 61.4  & 68.3 \\
    GPT-4o & 63.8  & 64.7  & -   & -   & -   & -   & 50.8  & 59.8  & 50.3  & 48.0  & 46.5 \\
    Qwen-VL-Plus & 43.3  & 35.5  & 31.2  & 54.6  & 48.1  & 51.4  & 21.3  & 26.0  & 21.2  & 18.5  & 19.1 \\
    \midrule
    \rowcolor[rgb]{ .851,  .882,  .949} \multicolumn{12}{c}{\textit{Open-Source MLLMs}} \\
    \midrule
    mPLUG-Owl2-7B & 22.2  & 23.6  & 10.2  & 22.7  & 27.2  & 27.9  & 10.3  & 11.6  & 11.4  & 11.1  & 9.4 \\
    MiniGPT4-7B & 23.1  & 26.0  & 13.4  & 18.6  & 30.4  & 30.2  & 12.2  & 12.3  & 12.9  & 12.5  & 14.8 \\
    LLaVA-1.5-13B & 27.7  & 22.7  & 18.9  & 23.8  & 43.0  & 30.2  & 12.7  & 17.1  & 12.0  & 12.6  & 12.7 \\
    SPHINX-V2-13B & 36.7  & 16.4  & 23.1  & 54.6  & 41.8  & 43.0  & 16.1  & 20.8  & 14.1  & 35.2  & 28.9 \\
    Deepseek-VL & 34.9  & 28.4  & 55.9  & 26.8  & 32.9  & 34.6  & 19.3  & 23.0  & 23.2  & 20.2  & 18.4 \\
    LLaVA-OneVision-7B & 51.1  & 66.8  & 54.8  & 39.4  & 57.6  & 40.8  & 20.7  &  25.5     &  21.8     &  20.9     & 21.2 \\
    Qwen2-VL-7B & 52.5  & 44.7  & 59.7  & 55.4  & 52.5  & 49.7  & 20.1  & 23.7  & 16.3  & 19.0  & 19.8 \\
    llama-3.2-11B-Vision & 46.5  & 38.0  & 51.6  & 50.9  & 51.9  & 39.7  & 17.8  & 19.2  & 17.9  & 15.6  & 15.5 \\
    G-LLaVA-7B & 25.1  & 48.7  & 3.6   & 19.1  & 25.0  & 28.7  & 16.6  & 20.9  & 20.7  & 17.2  & 14.6 \\
    Math-LLaVA-13B & 46.6  & 57.7  & 56.5  & 37.2  & 51.3  & 33.5  & 22.9  & 27.3  & 24.9  & 24.5  & 21.7 \\
    \midrule
    \textbf{Ours} & \textbf{65.3} & 60.1  & \textbf{73.1} & \textbf{68.8} & \textbf{67.7} & \textbf{55.9} & \textbf{25.7} & \textbf{30.2} & \textbf{26.3} & 25.3  & 23.2 \\
    \bottomrule
    \end{tabular}%
    }
  \label{tab:main_exp}%
  \vspace{-2mm}
\end{table*}%

\begin{figure}[t]
\centering
\includegraphics[width=0.47\textwidth]{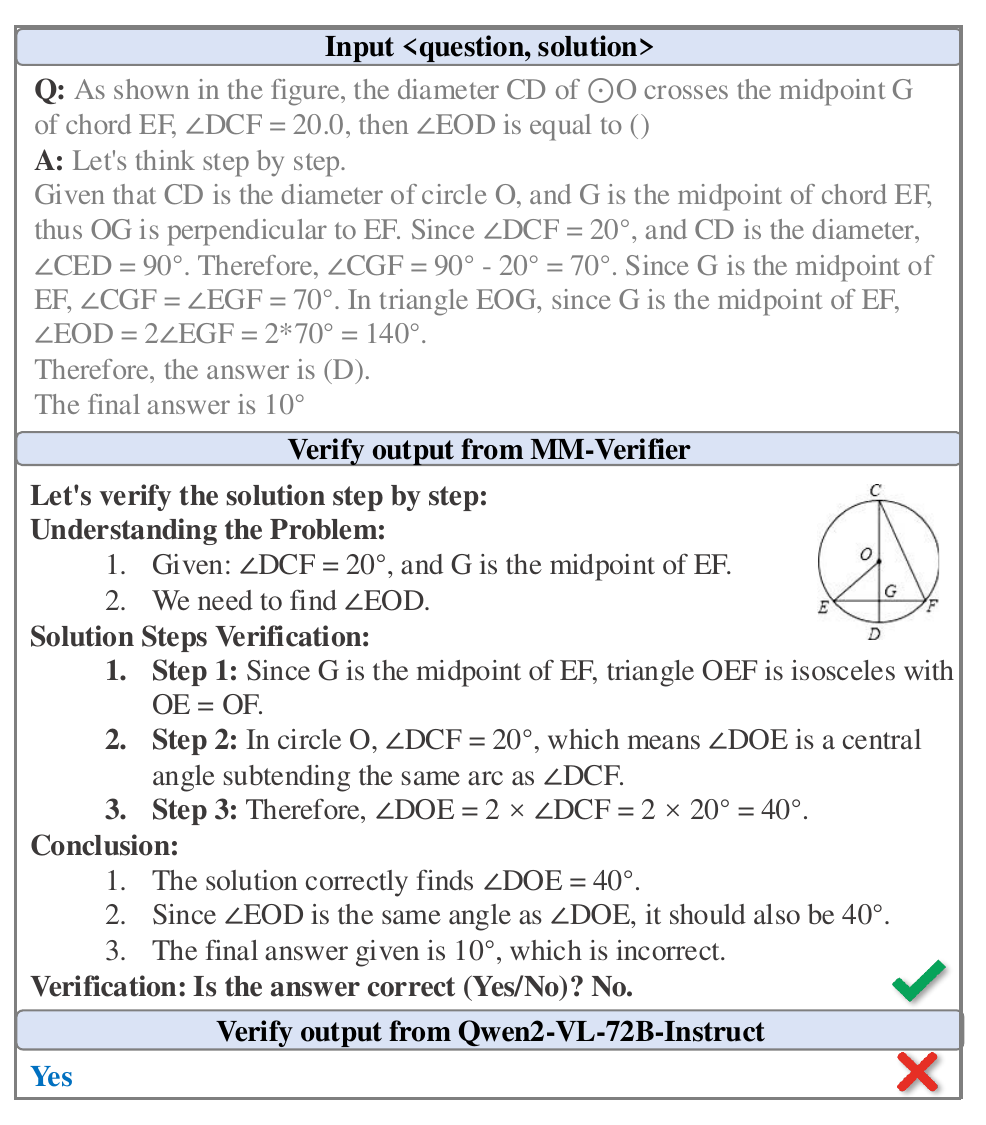}
\caption{We present a case of MM-Verifier. We can see MM-Verifier correctly verify the answer with Long COT while Qwen2-VL-72B-Instruct failed to.}
\label{fig:case}
\vspace{-7mm}
\end{figure}
\subsection{MM-Verifier and MM-Reasoner achieved SOTA Performance}
We leveraged our proposed MM-Reasoner and MM-Verifier together to enhance multimodal mathematical reasoning. Specifically, MM-Reasoner generated 12 diverse reasoning rollouts per query, while MM-Verifier systematically evaluated and filtered these rollouts, ensuring high-quality and accurate responses. This iterative verification-refinement process significantly improved reasoning precision and robustness.

As shown in Table~\ref{tab:main_exp}, our approach outperforms both open-source and closed-source MLLMs across multiple benchmarks. On the MATHVISTA dataset, our method achieves an overall accuracy of \textbf{65.3}, surpassing human performance (\textbf{60.3}) and even GPT-4o (\textbf{63.8}). Similarly, in the MATHVERSE benchmark, our method consistently achieves strong performance with an overall score of \textbf{25.7}, outperforming strong baselines like Math-LLaVA-13B (\textbf{22.9}) and LLaVA-OneVision (\textbf{20.7}). 

These results demonstrate the robustness of MM-Reasoner and MM-Verifier in improving mathematical reasoning across diverse tasks.

\subsection{Case Study}

Figure~\ref{fig:case} presents a case study of MM-Verifier, demonstrating its ability to successfully identify logical errors in the reasoning process through step-by-step verification. In contrast, Qwen2-VL-72B-Instruct fails to provide a step-by-step reasoning trajectory, leading to an error in detecting these mistakes. This case underscores the superior analytical capabilities of MM-Verifier.

\section{Conclusion}
In this paper, we propose two data synthesis methods: the first generates long COT verification data, while the second synthesizes long COT inference data more efficiently. We use these synthetic data to train the MM-Verifier and MM-Reasoner. Our MM-Verifier not only outperforms larger models on the MathCheck benchmark but also demonstrates superior performance against larger models on benchmarks such as MathVista and MathVerse. Additionally, the MM-Reasoner exhibits strong scalability, with performance improving as the data size increases. Furthermore, the combination of MM-Verifier and MM-Reasoner achieves impressive results on the MathVista benchmark, surpassing even GPT-4o. These findings confirm the effectiveness of MM-Verifier and MM-Reasoner in enhancing multimodal reasoning tasks and lay the foundation for future advancements in this domain.

\section{Limitations}
Due to limited funding and computational resources, we were unable to scale our MM-Verifier and MM-Reasoner to Qwen2-VL-72B. Additionally, our scalability tests were restricted to datasets of fewer than 100K samples. We plan to conduct further experiments as soon as additional computational resources become available.

\bibliography{custom}

\begin{thebibliography}{60}
\providecommand{\natexlab}[1]{#1}

\bibitem[{Bai et~al.(2023{\natexlab{a}})Bai, Bai, Chu, Cui, Dang, Deng, Fan, Ge, Han, Huang et~al.}]{bai2023qwen}
Jinze Bai, Shuai Bai, Yunfei Chu, Zeyu Cui, Kai Dang, Xiaodong Deng, Yang Fan, Wenbin Ge, Yu~Han, Fei Huang, et~al. 2023{\natexlab{a}}.
\newblock Qwen technical report.
\newblock \emph{arXiv preprint arXiv:2309.16609}.

\bibitem[{Bai et~al.(2023{\natexlab{b}})Bai, Bai, Yang, Wang, Tan, Wang, Lin, Zhou, and Zhou}]{bai2023qwenvl}
Jinze Bai, Shuai Bai, Shusheng Yang, Shijie Wang, Sinan Tan, Peng Wang, Junyang Lin, Chang Zhou, and Jingren Zhou. 2023{\natexlab{b}}.
\newblock Qwen-vl: A versatile vision-language model for understanding, localization, text reading, and beyond.

\bibitem[{Bai et~al.(2024)Bai, Liang, Wan, Yang, Li, Wang, Cui, He, Yuan, and Zhang}]{bai2024survey}
Tianyi Bai, Hao Liang, Binwang Wan, Ling Yang, Bozhou Li, Yifan Wang, Bin Cui, Conghui He, Binhang Yuan, and Wentao Zhang. 2024.
\newblock A survey of multimodal large language model from a data-centric perspective.
\newblock \emph{arXiv preprint arXiv:2405.16640}.

\bibitem[{Chen et~al.(2023)Chen, Li, Dong, Zhang, He, Wang, Zhao, and Lin}]{sharegpt4v}
Lin Chen, Jinsong Li, Xiaoyi Dong, Pan Zhang, Conghui He, Jiaqi Wang, Feng Zhao, and Dahua Lin. 2023.
\newblock Sharegpt4v: Improving large multi-modal models with better captions.
\newblock \emph{CoRR}, abs/2311.12793.

\bibitem[{Chen et~al.(2024)Chen, Wu, Wang, Su, Chen, Xing, Zhong, Zhang, Zhu, Lu et~al.}]{chen2024internvl}
Zhe Chen, Jiannan Wu, Wenhai Wang, Weijie Su, Guo Chen, Sen Xing, Muyan Zhong, Qinglong Zhang, Xizhou Zhu, Lewei Lu, et~al. 2024.
\newblock Internvl: Scaling up vision foundation models and aligning for generic visual-linguistic tasks.
\newblock In \emph{Proceedings of the IEEE/CVF Conference on Computer Vision and Pattern Recognition}, pages 24185--24198.

\bibitem[{Cui et~al.(2025)Cui, Yuan, Wang, Wang, Li, He, Fan, Yu, Xu, Chen et~al.}]{cui2025process}
Ganqu Cui, Lifan Yuan, Zefan Wang, Hanbin Wang, Wendi Li, Bingxiang He, Yuchen Fan, Tianyu Yu, Qixin Xu, Weize Chen, et~al. 2025.
\newblock Process reinforcement through implicit rewards.
\newblock \emph{arXiv preprint arXiv:2502.01456}.

\bibitem[{Du et~al.(2025)Du, Liu, Li, Zhao, Huo, Wang, Chen, Liu, Wang, and Wen}]{du2025virgo}
Yifan Du, Zikang Liu, Yifan Li, Wayne~Xin Zhao, Yuqi Huo, Bingning Wang, Weipeng Chen, Zheng Liu, Zhongyuan Wang, and Ji-Rong Wen. 2025.
\newblock Virgo: A preliminary exploration on reproducing o1-like mllm.
\newblock \emph{arXiv preprint arXiv:2501.01904}.

\bibitem[{Gao et~al.(2024)Gao, Cai, Xu, Wang, Zheng, Lin, Lu, Lin, Zhou, Xiao et~al.}]{gao2024llm}
Bofei Gao, Zefan Cai, Runxin Xu, Peiyi Wang, Ce~Zheng, Runji Lin, Keming Lu, Junyang Lin, Chang Zhou, Wen Xiao, et~al. 2024.
\newblock Llm critics help catch bugs in mathematics: Towards a better mathematical verifier with natural language feedback.
\newblock \emph{CoRR}.

\bibitem[{Gao et~al.(2023)Gao, Pi, Zhang, Ye, Zhong, Wang, Hong, Han, Xu, Li et~al.}]{gao2023g}
Jiahui Gao, Renjie Pi, Jipeng Zhang, Jiacheng Ye, Wanjun Zhong, Yufei Wang, Lanqing Hong, Jianhua Han, Hang Xu, Zhenguo Li, et~al. 2023.
\newblock G-llava: Solving geometric problem with multi-modal large language model.
\newblock \emph{arXiv preprint arXiv:2312.11370}.

\bibitem[{Gu et~al.(2024)Gu, Jiang, Shi, Tan, Zhai, Xu, Li, Shen, Ma, Liu et~al.}]{gu2024survey}
Jiawei Gu, Xuhui Jiang, Zhichao Shi, Hexiang Tan, Xuehao Zhai, Chengjin Xu, Wei Li, Yinghan Shen, Shengjie Ma, Honghao Liu, et~al. 2024.
\newblock A survey on llm-as-a-judge.
\newblock \emph{arXiv preprint arXiv:2411.15594}.

\bibitem[{Guo et~al.(2025)Guo, Yang, Zhang, Song, Zhang, Xu, Zhu, Ma, Wang, Bi et~al.}]{guo2025deepseek}
Daya Guo, Dejian Yang, Haowei Zhang, Junxiao Song, Ruoyu Zhang, Runxin Xu, Qihao Zhu, Shirong Ma, Peiyi Wang, Xiao Bi, et~al. 2025.
\newblock Deepseek-r1: Incentivizing reasoning capability in llms via reinforcement learning.
\newblock \emph{arXiv preprint arXiv:2501.12948}.

\bibitem[{Huang et~al.(2024)Huang, Yu, Xiong, He, Tang, and Fu}]{huang2024hologram}
Litian Huang, Xinguo Yu, Feng Xiong, Bin He, Shengbing Tang, and Jiawen Fu. 2024.
\newblock Hologram reasoning for solving algebra problems with geometry diagrams.
\newblock \emph{arXiv preprint arXiv:2408.10592}.

\bibitem[{Jian et~al.(2023)Jian, Gao, and Vosoughi}]{pformer}
Yiren Jian, Chongyang Gao, and Soroush Vosoughi. 2023.
\newblock Bootstrapping vision-language learning with decoupled language pre-training.
\newblock In \emph{Advances in Neural Information Processing Systems 36: Annual Conference on Neural Information Processing Systems 2023, NeurIPS 2023, New Orleans, LA, USA, December 10 - 16, 2023}.

\bibitem[{Li et~al.(2023{\natexlab{a}})Li, Zhang, Chen, Wang, Yang, and Liu}]{otter}
Bo~Li, Yuanhan Zhang, Liangyu Chen, Jinghao Wang, Jingkang Yang, and Ziwei Liu. 2023{\natexlab{a}}.
\newblock Otter: {A} multi-modal model with in-context instruction tuning.
\newblock \emph{CoRR}, abs/2305.03726.

\bibitem[{Li et~al.(2024{\natexlab{a}})Li, Zhang, Guo, Zhang, Li, Zhang, Zhang, Zhang, Li, Liu et~al.}]{li2024llava}
Bo~Li, Yuanhan Zhang, Dong Guo, Renrui Zhang, Feng Li, Hao Zhang, Kaichen Zhang, Peiyuan Zhang, Yanwei Li, Ziwei Liu, et~al. 2024{\natexlab{a}}.
\newblock Llava-onevision: Easy visual task transfer.
\newblock \emph{arXiv preprint arXiv:2408.03326}.

\bibitem[{Li et~al.(2023{\natexlab{b}})Li, Li, Savarese, and Hoi}]{blip}
Junnan Li, Dongxu Li, Silvio Savarese, and Steven Hoi. 2023{\natexlab{b}}.
\newblock Blip-2: Bootstrapping language-image pre-training with frozen image encoders and large language models.
\newblock In \emph{International conference on machine learning}, pages 19730--19742. PMLR.

\bibitem[{Li et~al.(2023{\natexlab{c}})Li, Li, Savarese, and Hoi}]{li2023blip}
Junnan Li, Dongxu Li, Silvio Savarese, and Steven Hoi. 2023{\natexlab{c}}.
\newblock Blip-2: Bootstrapping language-image pre-training with frozen image encoders and large language models.
\newblock In \emph{International conference on machine learning}, pages 19730--19742. PMLR.

\bibitem[{Li et~al.(2022{\natexlab{a}})Li, Li, Xiong, and Hoi}]{blip1}
Junnan Li, Dongxu Li, Caiming Xiong, and Steven C.~H. Hoi. 2022{\natexlab{a}}.
\newblock {BLIP:} bootstrapping language-image pre-training for unified vision-language understanding and generation.
\newblock In \emph{International Conference on Machine Learning, {ICML} 2022, 17-23 July 2022, Baltimore, Maryland, {USA}}, volume 162, pages 12888--12900.

\bibitem[{Li et~al.(2022{\natexlab{b}})Li, Zhang, Zhang, Yang, Li, Zhong, Wang, Yuan, Zhang, Hwang, Chang, and Gao}]{grounded-pt}
Liunian~Harold Li, Pengchuan Zhang, Haotian Zhang, Jianwei Yang, Chunyuan Li, Yiwu Zhong, Lijuan Wang, Lu~Yuan, Lei Zhang, Jenq{-}Neng Hwang, Kai{-}Wei Chang, and Jianfeng Gao. 2022{\natexlab{b}}.
\newblock Grounded language-image pre-training.
\newblock In \emph{{IEEE/CVF} Conference on Computer Vision and Pattern Recognition, {CVPR} 2022, New Orleans, LA, USA, June 18-24, 2022}, pages 10955--10965. {IEEE}.

\bibitem[{Li et~al.(2024{\natexlab{b}})Li, Du, Liu, Zhang, Liu, Zhang, and Cai}]{li2024eagle}
Zhihao Li, Yao Du, Yang Liu, Yan Zhang, Yufang Liu, Mengdi Zhang, and Xunliang Cai. 2024{\natexlab{b}}.
\newblock Eagle: Elevating geometric reasoning through llm-empowered visual instruction tuning.
\newblock \emph{arXiv preprint arXiv:2408.11397}.

\bibitem[{Liang et~al.(2023)Liang, Yang, Zhang, and Zhang}]{liang2023unimath}
Zhenwen Liang, Tianyu Yang, Jipeng Zhang, and Xiangliang Zhang. 2023.
\newblock Unimath: A foundational and multimodal mathematical reasoner.
\newblock In \emph{Proceedings of the 2023 Conference on Empirical Methods in Natural Language Processing}, pages 7126--7133.

\bibitem[{Lin et~al.(2023)Lin, Liu, Zhang, Gao, Qiu, Xiao, Qiu, Lin, Shao, Chen et~al.}]{lin2023sphinx}
Ziyi Lin, Chris Liu, Renrui Zhang, Peng Gao, Longtian Qiu, Han Xiao, Han Qiu, Chen Lin, Wenqi Shao, Keqin Chen, et~al. 2023.
\newblock Sphinx: The joint mixing of weights, tasks, and visual embeddings for multi-modal large language models.
\newblock \emph{arXiv preprint arXiv:2311.07575}.

\bibitem[{Liu et~al.(2023{\natexlab{a}})Liu, Li, Li, and Lee}]{llava1.5}
Haotian Liu, Chunyuan Li, Yuheng Li, and Yong~Jae Lee. 2023{\natexlab{a}}.
\newblock Improved baselines with visual instruction tuning.
\newblock \emph{arXiv preprint arXiv:2310.03744}.

\bibitem[{Liu et~al.(2023{\natexlab{b}})Liu, Li, Wu, and Lee}]{llava}
Haotian Liu, Chunyuan Li, Qingyang Wu, and Yong~Jae Lee. 2023{\natexlab{b}}.
\newblock Visual instruction tuning.
\newblock In \emph{Advances in Neural Information Processing Systems 36: Annual Conference on Neural Information Processing Systems 2023, NeurIPS 2023, New Orleans, LA, USA, December 10 - 16, 2023}.

\bibitem[{Liu et~al.(2023{\natexlab{c}})Liu, Zeng, Ren, Li, Zhang, Yang, Li, Yang, Su, Zhu, and Zhang}]{dino}
Shilong Liu, Zhaoyang Zeng, Tianhe Ren, Feng Li, Hao Zhang, Jie Yang, Chunyuan Li, Jianwei Yang, Hang Su, Jun Zhu, and Lei Zhang. 2023{\natexlab{c}}.
\newblock Grounding {DINO:} marrying {DINO} with grounded pre-training for open-set object detection.
\newblock \emph{CoRR}, abs/2303.05499.

\bibitem[{Liu et~al.(2024)Liu, Li, Zhang, Zhou, Cheng, and He}]{liu2024diving}
Wei Liu, Junlong Li, Xiwen Zhang, Fan Zhou, Yu~Cheng, and Junxian He. 2024.
\newblock Diving into self-evolving training for multimodal reasoning.
\newblock \emph{arXiv preprint arXiv:2412.17451}.

\bibitem[{Lu et~al.(2024)Lu, Liu, Zhang, Wang, Dong, Liu, Sun, Ren, Li, Sun et~al.}]{lu2024deepseek}
Haoyu Lu, Wen Liu, Bo~Zhang, Bingxuan Wang, Kai Dong, Bo~Liu, Jingxiang Sun, Tongzheng Ren, Zhuoshu Li, Yaofeng Sun, et~al. 2024.
\newblock Deepseek-vl: towards real-world vision-language understanding.
\newblock \emph{arXiv preprint arXiv:2403.05525}.

\bibitem[{Lu et~al.(2023{\natexlab{a}})Lu, Gan, Zhang, Wu, Wu, Sun, Zhang, Zhang, and Song}]{lyrics}
Junyu Lu, Ruyi Gan, Dixiang Zhang, Xiaojun Wu, Ziwei Wu, Renliang Sun, Jiaxing Zhang, Pingjian Zhang, and Yan Song. 2023{\natexlab{a}}.
\newblock Lyrics: Boosting fine-grained language-vision alignment and comprehension via semantic-aware visual objects.
\newblock \emph{CoRR}, abs/2312.05278.

\bibitem[{Lu et~al.(2023{\natexlab{b}})Lu, Bansal, Xia, Liu, Li, Hajishirzi, Cheng, Chang, Galley, and Gao}]{lu2023mathvista}
Pan Lu, Hritik Bansal, Tony Xia, Jiacheng Liu, Chunyuan Li, Hannaneh Hajishirzi, Hao Cheng, Kai-Wei Chang, Michel Galley, and Jianfeng Gao. 2023{\natexlab{b}}.
\newblock Mathvista: Evaluating mathematical reasoning of foundation models in visual contexts.
\newblock \emph{arXiv preprint arXiv:2310.02255}.

\bibitem[{Luo et~al.(2025)Luo, Zheng, Wang, Yu, Ni, Lin, Zeng, and Yang}]{luo2025ursa}
Ruilin Luo, Zhuofan Zheng, Yifan Wang, Yiyao Yu, Xinzhe Ni, Zicheng Lin, Jin Zeng, and Yujiu Yang. 2025.
\newblock Ursa: Understanding and verifying chain-of-thought reasoning in multimodal mathematics.
\newblock \emph{arXiv preprint arXiv:2501.04686}.

\bibitem[{Meidani et~al.(2023)Meidani, Shojaee, Reddy, and Farimani}]{meidani2023snip}
Kazem Meidani, Parshin Shojaee, Chandan~K Reddy, and Amir~Barati Farimani. 2023.
\newblock Snip: Bridging mathematical symbolic and numeric realms with unified pre-training.
\newblock \emph{arXiv preprint arXiv:2310.02227}.

\bibitem[{Min et~al.(2024)Min, Chen, Jiang, Chen, Deng, Hu, Tang, Wang, Cheng, Song et~al.}]{min2024imitate}
Yingqian Min, Zhipeng Chen, Jinhao Jiang, Jie Chen, Jia Deng, Yiwen Hu, Yiru Tang, Jiapeng Wang, Xiaoxue Cheng, Huatong Song, et~al. 2024.
\newblock Imitate, explore, and self-improve: A reproduction report on slow-thinking reasoning systems.
\newblock \emph{arXiv preprint arXiv:2412.09413}.

\bibitem[{Muennighoff et~al.(2025)Muennighoff, Yang, Shi, Li, Fei-Fei, Hajishirzi, Zettlemoyer, Liang, Cand{\`e}s, and Hashimoto}]{muennighoff2025s1}
Niklas Muennighoff, Zitong Yang, Weijia Shi, Xiang~Lisa Li, Li~Fei-Fei, Hannaneh Hajishirzi, Luke Zettlemoyer, Percy Liang, Emmanuel Cand{\`e}s, and Tatsunori Hashimoto. 2025.
\newblock s1: Simple test-time scaling.
\newblock \emph{arXiv preprint arXiv:2501.19393}.

\bibitem[{o1~Team(2024)}]{skyworkopeno12024}
Skywork o1~Team. 2024.
\newblock \href {https://huggingface.co/Skywork} {Skywork-o1 open series}.
\newblock \url{https://huggingface.co/Skywork}.

\bibitem[{OpenAI(2023{\natexlab{a}})}]{chatgpt}
OpenAI. 2023{\natexlab{a}}.
\newblock \href {https://openai.com/blog/chatgpt} {Chatgpt}.

\bibitem[{OpenAI(2023{\natexlab{b}})}]{openai2023gpt}
R~OpenAI. 2023{\natexlab{b}}.
\newblock Gpt-4 technical report. arxiv 2303.08774.
\newblock \emph{View in Article}, 2(5).

\bibitem[{Radford et~al.(2021)Radford, Kim, Hallacy, Ramesh, Goh, Agarwal, Sastry, Askell, Mishkin, Clark et~al.}]{clip}
Alec Radford, Jong~Wook Kim, Chris Hallacy, Aditya Ramesh, Gabriel Goh, Sandhini Agarwal, Girish Sastry, Amanda Askell, Pamela Mishkin, Jack Clark, et~al. 2021.
\newblock Learning transferable visual models from natural language supervision.
\newblock In \emph{International conference on machine learning}, pages 8748--8763. PMLR.

\bibitem[{Shi et~al.(2024)Shi, Hu, Bin, Liu, Yang, Ng, Bing, and Lee}]{shi2024math}
Wenhao Shi, Zhiqiang Hu, Yi~Bin, Junhua Liu, Yang Yang, See-Kiong Ng, Lidong Bing, and Roy Ka-Wei Lee. 2024.
\newblock Math-llava: Bootstrapping mathematical reasoning for multimodal large language models.
\newblock \emph{arXiv preprint arXiv:2406.17294}.

\bibitem[{Sun et~al.(2024)Sun, Liang, Wei, Yu, He, Zhou, and Zhang}]{sun2024beats}
Linzhuang Sun, Hao Liang, Jingxuan Wei, Bihui Yu, Conghui He, Zenan Zhou, and Wentao Zhang. 2024.
\newblock Beats: Optimizing llm mathematical capabilities with backverify and adaptive disambiguate based efficient tree search.
\newblock \emph{arXiv preprint arXiv:2409.17972}.

\bibitem[{Team(2024)}]{qwq-32b-preview}
Qwen Team. 2024.
\newblock \href {https://qwenlm.github.io/blog/qwq-32b-preview/} {Qwq: Reflect deeply on the boundaries of the unknown}.

\bibitem[{Thawakar et~al.(2025)Thawakar, Dissanayake, More, Thawkar, Heakl, Ahsan, Li, Zumri, Lahoud, Anwer et~al.}]{thawakar2025llamav}
Omkar Thawakar, Dinura Dissanayake, Ketan More, Ritesh Thawkar, Ahmed Heakl, Noor Ahsan, Yuhao Li, Mohammed Zumri, Jean Lahoud, Rao~Muhammad Anwer, et~al. 2025.
\newblock Llamav-o1: Rethinking step-by-step visual reasoning in llms.
\newblock \emph{arXiv preprint arXiv:2501.06186}.

\bibitem[{Touvron et~al.(2023)Touvron, Lavril, Izacard, Martinet, Lachaux, Lacroix, Rozi{\`e}re, Goyal, Hambro, Azhar et~al.}]{llama}
Hugo Touvron, Thibaut Lavril, Gautier Izacard, Xavier Martinet, Marie-Anne Lachaux, Timoth{\'e}e Lacroix, Baptiste Rozi{\`e}re, Naman Goyal, Eric Hambro, Faisal Azhar, et~al. 2023.
\newblock Llama: Open and efficient foundation language models.
\newblock \emph{arXiv preprint arXiv:2302.13971}.

\bibitem[{Wang et~al.(2024{\natexlab{a}})Wang, Li, Shao, Xu, Dai, Li, Chen, Wu, and Sui}]{wang2024math}
Peiyi Wang, Lei Li, Zhihong Shao, Runxin Xu, Damai Dai, Yifei Li, Deli Chen, Yu~Wu, and Zhifang Sui. 2024{\natexlab{a}}.
\newblock Math-shepherd: Verify and reinforce llms step-by-step without human annotations.
\newblock In \emph{Proceedings of the 62nd Annual Meeting of the Association for Computational Linguistics (Volume 1: Long Papers)}, pages 9426--9439.

\bibitem[{Wang et~al.(2024{\natexlab{b}})Wang, Mrini, Yang, Kumar, Tian, Yan, and Wang}]{image-text-data}
Weizhi Wang, Khalil Mrini, Linjie Yang, Sateesh Kumar, Yu~Tian, Xifeng Yan, and Heng Wang. 2024{\natexlab{b}}.
\newblock Finetuned multimodal language models are high-quality image-text data filters.
\newblock \emph{CoRR}, abs/2403.02677.

\bibitem[{Weng et~al.(2022)Weng, Zhu, Xia, Li, He, Liu, Sun, Liu, and Zhao}]{weng2022large}
Yixuan Weng, Minjun Zhu, Fei Xia, Bin Li, Shizhu He, Shengping Liu, Bin Sun, Kang Liu, and Jun Zhao. 2022.
\newblock Large language models are better reasoners with self-verification.
\newblock \emph{arXiv preprint arXiv:2212.09561}.

\bibitem[{Wu et~al.(2023)Wu, Gan, Chen, Wan, and Yu}]{wu2023multimodal}
Jiayang Wu, Wensheng Gan, Zefeng Chen, Shicheng Wan, and Philip~S Yu. 2023.
\newblock Multimodal large language models: A survey.
\newblock \emph{arXiv preprint arXiv:2311.13165}.

\bibitem[{Xiong et~al.(2024)Xiong, Zhang, Jiang, and Zhang}]{wei2024implementation}
Wei Xiong, Hanning Zhang, Nan Jiang, and Tong Zhang. 2024.
\newblock \href {https://github.com/RLHFlow/RLHF-Reward-Modeling} {An implementation of generative prm}.
\newblock \url{https://github.com/RLHFlow/RLHF-Reward-Modeling}.

\bibitem[{Ye et~al.(2024)Ye, Xu, Ye, Yan, Hu, Liu, Qian, Zhang, and Huang}]{ye2024mplug}
Qinghao Ye, Haiyang Xu, Jiabo Ye, Ming Yan, Anwen Hu, Haowei Liu, Qi~Qian, Ji~Zhang, and Fei Huang. 2024.
\newblock mplug-owl2: Revolutionizing multi-modal large language model with modality collaboration.
\newblock In \emph{Proceedings of the IEEE/CVF Conference on Computer Vision and Pattern Recognition}, pages 13040--13051.

\bibitem[{Ye et~al.(2025)Ye, Huang, Xiao, Chern, Xia, and Liu}]{ye2025limo}
Yixin Ye, Zhen Huang, Yang Xiao, Ethan Chern, Shijie Xia, and Pengfei Liu. 2025.
\newblock Limo: Less is more for reasoning.
\newblock \emph{arXiv preprint arXiv:2502.03387}.

\bibitem[{Zhang et~al.(2024{\natexlab{a}})Zhang, Huang, Zhou, Li, and Ouyang}]{zhang2024accessing}
Di~Zhang, Xiaoshui Huang, Dongzhan Zhou, Yuqiang Li, and Wanli Ouyang. 2024{\natexlab{a}}.
\newblock Accessing gpt-4 level mathematical olympiad solutions via monte carlo tree self-refine with llama-3 8b.
\newblock \emph{arXiv preprint arXiv:2406.07394}.

\bibitem[{Zhang et~al.(2024{\natexlab{b}})Zhang, Lei, Li, Wang, Liu, Yang, Li, Wang, Yang, Wu et~al.}]{zhang2024critic}
Di~Zhang, Jingdi Lei, Junxian Li, Xunzhi Wang, Yujie Liu, Zonglin Yang, Jiatong Li, Weida Wang, Suorong Yang, Jianbo Wu, et~al. 2024{\natexlab{b}}.
\newblock Critic-v: Vlm critics help catch vlm errors in multimodal reasoning.
\newblock \emph{arXiv preprint arXiv:2411.18203}.

\bibitem[{Zhang et~al.(2022)Zhang, Zhang, Hu, Chen, Li, Dai, Wang, Yuan, Hwang, and Gao}]{glipv2}
Haotian Zhang, Pengchuan Zhang, Xiaowei Hu, Yen{-}Chun Chen, Liunian~Harold Li, Xiyang Dai, Lijuan Wang, Lu~Yuan, Jenq{-}Neng Hwang, and Jianfeng Gao. 2022.
\newblock Glipv2: Unifying localization and vision-language understanding.
\newblock In \emph{Advances in Neural Information Processing Systems 35: Annual Conference on Neural Information Processing Systems 2022, NeurIPS 2022, New Orleans, LA, USA, November 28 - December 9, 2022}.

\bibitem[{Zhang et~al.(2024{\natexlab{c}})Zhang, Hosseini, Bansal, Kazemi, Kumar, and Agarwal}]{zhang2024generative}
Lunjun Zhang, Arian Hosseini, Hritik Bansal, Mehran Kazemi, Aviral Kumar, and Rishabh Agarwal. 2024{\natexlab{c}}.
\newblock Generative verifiers: Reward modeling as next-token prediction.
\newblock \emph{arXiv preprint arXiv:2408.15240}.

\bibitem[{Zhang et~al.(2024{\natexlab{d}})Zhang, Jiang, Zhang, Lin, Guo, Qiu, Zhou, Lu, Chang, Gao et~al.}]{zhang2024mathverse}
Renrui Zhang, Dongzhi Jiang, Yichi Zhang, Haokun Lin, Ziyu Guo, Pengshuo Qiu, Aojun Zhou, Pan Lu, Kai-Wei Chang, Peng Gao, et~al. 2024{\natexlab{d}}.
\newblock Mathverse: Does your multi-modal llm truly see the diagrams in visual math problems?
\newblock \emph{arXiv preprint arXiv:2403.14624}.

\bibitem[{Zhang et~al.(2024{\natexlab{e}})Zhang, Wei, Jiang, Zhang, Guo, Tong, Liu, Zhou, Wei, Zhang et~al.}]{zhang2024mavis}
Renrui Zhang, Xinyu Wei, Dongzhi Jiang, Yichi Zhang, Ziyu Guo, Chengzhuo Tong, Jiaming Liu, Aojun Zhou, Bin Wei, Shanghang Zhang, et~al. 2024{\natexlab{e}}.
\newblock Mavis: Mathematical visual instruction tuning.
\newblock \emph{arXiv preprint arXiv:2407.08739}.

\bibitem[{Zhang et~al.(2025)Zhang, Zheng, Wu, Zhang, Lin, Yu, Liu, Zhou, and Lin}]{zhang2025lessons}
Zhenru Zhang, Chujie Zheng, Yangzhen Wu, Beichen Zhang, Runji Lin, Bowen Yu, Dayiheng Liu, Jingren Zhou, and Junyang Lin. 2025.
\newblock The lessons of developing process reward models in mathematical reasoning.
\newblock \emph{arXiv preprint arXiv:2501.07301}.

\bibitem[{Zhao et~al.(2023)Zhao, Zhou, Li, Tang, Wang, Hou, Min, Zhang, Zhang, Dong et~al.}]{zhao2023survey}
Wayne~Xin Zhao, Kun Zhou, Junyi Li, Tianyi Tang, Xiaolei Wang, Yupeng Hou, Yingqian Min, Beichen Zhang, Junjie Zhang, Zican Dong, et~al. 2023.
\newblock A survey of large language models.
\newblock \emph{arXiv preprint arXiv:2303.18223}.

\bibitem[{Zhou et~al.(2024)Zhou, Liu, Ning, Liu, Wang, Wong, Huang, Wang, and Huang}]{zhou2024your}
Zihao Zhou, Shudong Liu, Maizhen Ning, Wei Liu, Jindong Wang, Derek~F Wong, Xiaowei Huang, Qiufeng Wang, and Kaizhu Huang. 2024.
\newblock Is your model really a good math reasoner? evaluating mathematical reasoning with checklist.
\newblock \emph{arXiv preprint arXiv:2407.08733}.

\bibitem[{Zhu et~al.(2023)Zhu, Chen, Shen, Li, and Elhoseiny}]{zhu2023minigpt}
Deyao Zhu, Jun Chen, Xiaoqian Shen, Xiang Li, and Mohamed Elhoseiny. 2023.
\newblock Minigpt-4: Enhancing vision-language understanding with advanced large language models.
\newblock \emph{arXiv preprint arXiv:2304.10592}.

\bibitem[{Zhuang et~al.(2024)Zhuang, Huang, Zhang, and Zeng}]{zhuang2024math}
Wenwen Zhuang, Xin Huang, Xiantao Zhang, and Jin Zeng. 2024.
\newblock Math-puma: Progressive upward multimodal alignment to enhance mathematical reasoning.
\newblock \emph{arXiv preprint arXiv:2408.08640}.

\end{thebibliography}

\clearpage
\appendix

\begin{table*}[htbp]
  \centering
  \caption{We compare sub-task's performance of Qwen2-VL-Instruct-7B, LLaMA-3.2-11B-Vision with our MM-Reasoner on the MathVista benchmark. GPS: geometry problem solving; MWP: math word problem; FQA: figure question answering; TQA: textbook question answering; VQA: visual question answering.}
  \resizebox{\linewidth}{!}{
    \begin{tabular}{cl|c|c|c|c|c|c}
    \toprule
    \textbf{Model} & \multicolumn{1}{c|}{\textbf{Method}} & \textbf{GPS} & \textbf{MWP} & \textbf{FQA} & \textbf{TQA} & \textbf{VQA} & \textbf{ALL} \\
    \midrule
    \rowcolor[rgb]{ .851,  .882,  .949} \multicolumn{8}{c}{\textit{Sample 4}} \\
    \midrule
    \multirow{4}[2]{*}{Qwen2-VL} & Majority Voting & 44.7  & 62.9  & 63.6  & 58.2  & 54.7  & 57.1 \\
          & Qwen2-VL-7B as Judgement & 51.4  & 63.4  & 62.8  & 53.8  & 55.9  & 57.9 \\
          & MM-Verifier(Stage1) & 51.0  & 65.1  & 62.5  & 63.3  & 52.0  & 58.8 \\
          & MM-Verifier(Stage2) & \textbf{51.9}  & \textbf{66.7}  & \textbf{64.7}  & 60.8  & 53.6  & \textbf{59.8} \\
    \midrule
    \multirow{4}[1]{*}{LLaMA-3.2-11B-Vision} & Majority Voting & 38.0  & 48.9  & 45.7  & 55.7  & 39.7  & 45.2 \\
          & Qwen2-VL-7B as Judgement & 39.9  & 51.6  & 48.7  & 52.5  & 38.0  & 41.7 \\
          & MM-Verifier(Stage1) & 42.3  & 54.8  & 51.7  & 57.0  & 45.3  & 50.0 \\
          & MM-Verifier(Stage2) & \textbf{46.2}  & 53.8  & \textbf{53.2}  & 56.3  & 42.5  & \textbf{50.4} \\
    \midrule
    \multirow{4}[1]{*}{MM-Reasoner} & Majority Voting & 57.2  & 62.4  & 65.4  & 57.6  & 51.4  & 59.4 \\
          & Qwen2-VL-7B as Judgement & 45.7  & 59.1  & 56.1  & 55.7  & 48.6  & 53.1 \\
          & MM-Verifier(Stage1) & 51.0  & 65.1  & 65.1  & 61.4  & 54.2  & 59.6 \\
          & MM-Verifier(Stage2) & \textbf{58.7}  & \textbf{66.1}  & \textbf{66.2}  & \textbf{63.9}  & 50.8  & \textbf{61.5} \\
    \midrule
    \rowcolor[rgb]{ .851,  .882,  .949} \multicolumn{8}{c}{\textit{Sample 8}} \\
    \midrule
    \multirow{4}[2]{*}{Qwen2-VL} & Majority Voting & 55.8  & 65.1  & 65.8  & 62.0  & 55.3  & 61.1 \\
          & Qwen2-VL-7B as Judgement & 49.0  & 60.2  & 59.1  & 52.5  & 49.7  & 54.5 \\
          & MM-Verifier(Stage1) & 53.8  & 69.9  & 66.2  & 63.9  & 53.1  & 61.6 \\
          & MM-Verifier(Stage2) & 52.4  & \textbf{70.4}  & \textbf{69.5}  & 62.7  & \textbf{55.3}  & \textbf{62.5} \\
    \midrule
    \multirow{4}[1]{*}{LLaMA-3.2-11B-Vision} & Majority Voting & 43.8  & 52.7  & 50.6  & 57.0  & 38.0  & 48.3 \\
          & Qwen2-VL-7B as Judgement & 39.9  & 51.6  & 48.7  & 52.5  & 38.0  & 46.1 \\
          & MM-Verifier(Stage1) & 45.2  & 54.3  & 54.6  & 62.0  & 41.3  & 51.4 \\
          & MM-Verifier(Stage2) & 44.7  & \textbf{55.4}  & \textbf{55.0}  & \textbf{63.9}  & \textbf{42.5}  & \textbf{52.1} \\
    \midrule
    \multirow{4}[1]{*}{MM-Reasoner} & Majority Voting & 57.2  & 66.7  & 66.2  & 64.6  & 55.3  & 62.2 \\
          & Qwen2-VL-7B as Judgement & 47.6  & 52.2  & 59.1  & 56.3  & 51.4  & 53.6 \\
          & MM-Verifier(Stage1) & 56.7  & 72.0  & 66.2  & 68.4  & 55.7  & 63.8 \\
          & MM-Verifier(Stage2) & \textbf{60.0}  & \textbf{73.1}  & \textbf{68.8}  & 67.7  & \textbf{55.9}  & \textbf{65.3} \\
    \midrule
    \rowcolor[rgb]{ .851,  .882,  .949} \multicolumn{8}{c}{\textit{Sample 12}} \\
    \midrule
    \multirow{4}[2]{*}{Qwen2-VL} & Majority Voting & 62.5  & 67.7  & 65.4  & 62.7  & 54.7  & 62.9 \\
          & Qwen2-VL-7B as Judgement & 49.5  & 60.8  & 59.5  & 51.3  & 48.6  & 54.4 \\
          & MM-Verifier(Stage1) & 56.7  & 69.9  & 69.9  & 65.2  & 54.7  & 63.7 \\
          & MM-Verifier(Stage2) & \textbf{58.7}  & \textbf{67.7}  & \textbf{69.5}  & \textbf{65.8}  & \textbf{57.0}  & \textbf{64.1} \\
    \midrule
    \multirow{4}[2]{*}{LLaMA-3.2-11B-Vision} & Majority Voting & 47.1  & 54.3  & 54.3  & 61.4  & 39.7  & 51.3 \\
          & Qwen2-VL-7B as Judgement & 42.3  & 45.2  & 50.6  & 54.4  & 34.1  & 45.5 \\
          & MM-Verifier(Stage1) & 50.5  & 57.5  & 58.0  & 65.8  & 43.6  & 55.0 \\
          & MM-Verifier(Stage2) & \textbf{55.8}  & 56.5  & \textbf{59.5}  & 62.7  & \textbf{44.1}  & \textbf{55.9} \\
    \midrule
    \multirow{4}[2]{*}{MM-Reasoner} & Majority Voting & 65.2  & 68.0  & 68.2  & 64.4  & 55.3  & 64.6 \\
          & Qwen2-VL-7B as Judgement & 46.6  & 61.3  & 59.5  & 56.3  & 52.5  & 55.4 \\
          & MM-Verifier(Stage1) & 61.1  & 72.0  & 70;3  & 64.6  & 53.1  & 64.8 \\
          & MM-Verifier(Stage2) & 62.0  & 71.5  & 69.9  & \textbf{67.1}  & 53.6  & \textbf{65.3} \\
    \bottomrule
    \end{tabular}%
    }
  \label{tab:appendix_vista}%
\end{table*}%

\begin{table*}[!t!]\footnotesize
  \centering
  \caption{Comparison of the performance of MCTS with Majority Voting and our MM-Verify on the MathVista benchmark.}
  \resizebox{1.0\linewidth}{!}{
    \begin{tabular}{lclcccccc}
    \toprule
    \multicolumn{1}{l}{\textbf{Base Model}} & \textbf{\# size} & \textbf{Method} & \textbf{GPS} & \textbf{MWP} & \textbf{FQA} & \textbf{TQA} & \textbf{VQA} & \textbf{ALL} \\
    \midrule
    \multirow{3}{*}{llama-3.2-11B-Vision} & \multirow{3}{*}{11B} & MCTS  & 42.8  & 16.1  & 33.8  & 48.1  & 40.2  & 35.8 \\
          &       & Majority Voting & 38.0  & 48.9  & 45.7  & 55.7  & 39.7  & 45.2 \\
          &       & \textbf{MM-Verify} & \textbf{46.2}  & \textbf{53.8}  & \textbf{53.2}  & \textbf{56.3}  & \textbf{42.5}  & \textbf{50.4} \\
    \midrule
    \multirow{3}{*}{Qwen2-VL} & \multirow{3}{*}{7B} & MCTS  & \textbf{53.4}  & 47.3  & 51.7  & 55.7  & 43.6  & 50.4 \\
          &       & Majority Voting & 44.7  & 62.9  & 63.6  & 58.2  & \textbf{54.7}  & 57.1 \\
          &       & \textbf{MM-Verify} & 51.9  & \textbf{66.7}  & \textbf{64.7}  & \textbf{60.8}  & 53.6  & \textbf{59.8} \\
    \midrule
    \multirow{3}{*}{LLaVA-OneVision} & \multirow{3}{*}{7B} & MCTS  & \textbf{72.1}  & 56.5  & 44.6  & \textbf{58.9}  & \textbf{45.3}  & 54.9 \\
          &       & Majority Voting & 66.8  & 54.8  & 39.4  & 57.6  & 40.8  & 51.1 \\
          &       & \textbf{MM-Verify} & 66.8  & \textbf{63.4}  & \textbf{47.6}  & 58.2  & 43.0  & \textbf{55.4} \\
    \bottomrule
    \end{tabular}%
    }
  \label{tab:mcts}%
\end{table*}%

\section{Performance of Naive MCTS in Multimodal Reasoning}~\label{app:MCTS_performance}

In NLP-based mathematical reasoning tasks, given a question, the Monte Carlo Tree Search (MCTS) algorithm iteratively refines its responses through continuous self-reflection, ultimately converging to an optimized answer~\cite{zhang2024accessing}. Experimental results demonstrate that this search-based approach yields strong performance. Naturally, we hypothesized that applying the same methodology to multimodal tasks would lead to similar improvements. However, as shown in Table~\ref{tab:mcts},
we compare the performance of MCTS with Majority Voting and our MM-Verifier on the MathVista benchmark. We can see simple MCTS failed to improve reasoning performance. We attribute this to the significantly higher prevalence of hallucinations in multimodal models compared to their text-only counterparts. Therefore inspired by~\cite{wang2024math, sun2024beats} we design a simulation-based tree search.

\section{Detailed Performance of MM-Verifier on Sub-tasks of MathVista}

Table~\ref{tab:appendix_vista} presents a comparative evaluation of Qwen2-VL-Instruct-7B, LLaMA-3.2-11B-Vision, and our proposed MM-Reasoner on the MathVista benchmark. The models were assessed across multiple reasoning-intensive sub-tasks, including GPS, MWP, FQA, TQA, and VQA. 

In the Sample 4 evaluation, MM-Reasoner with the MM-Verifier (Stage 2) achieved an overall accuracy of 61.5, surpassing Qwen2-VL's 59.8 and LLaMA-3.2-11B-Vision's 50.4. This trend persists across Sample 8 and Sample 12, where MM-Reasoner obtained 65.3 and 65.7 respectively, further establishing its robustness in complex multi-modal reasoning tasks. 

Furthermore, the MM-Verifier mechanism contributes significantly to the accuracy gains. Across all models, the transition from Majority Voting to MM-Verifier (Stage 2) consistently improves performance, underscoring the importance of verification-enhanced reasoning. In particular, MM-Reasoner benefits the most from this verification process, indicating that it effectively integrates verification feedback into its reasoning pipeline. Overall, these results validate the effectiveness of MM-Verify as a strong algorithm.

\begin{figure}[t]
\centering
\includegraphics[width=0.47\textwidth]{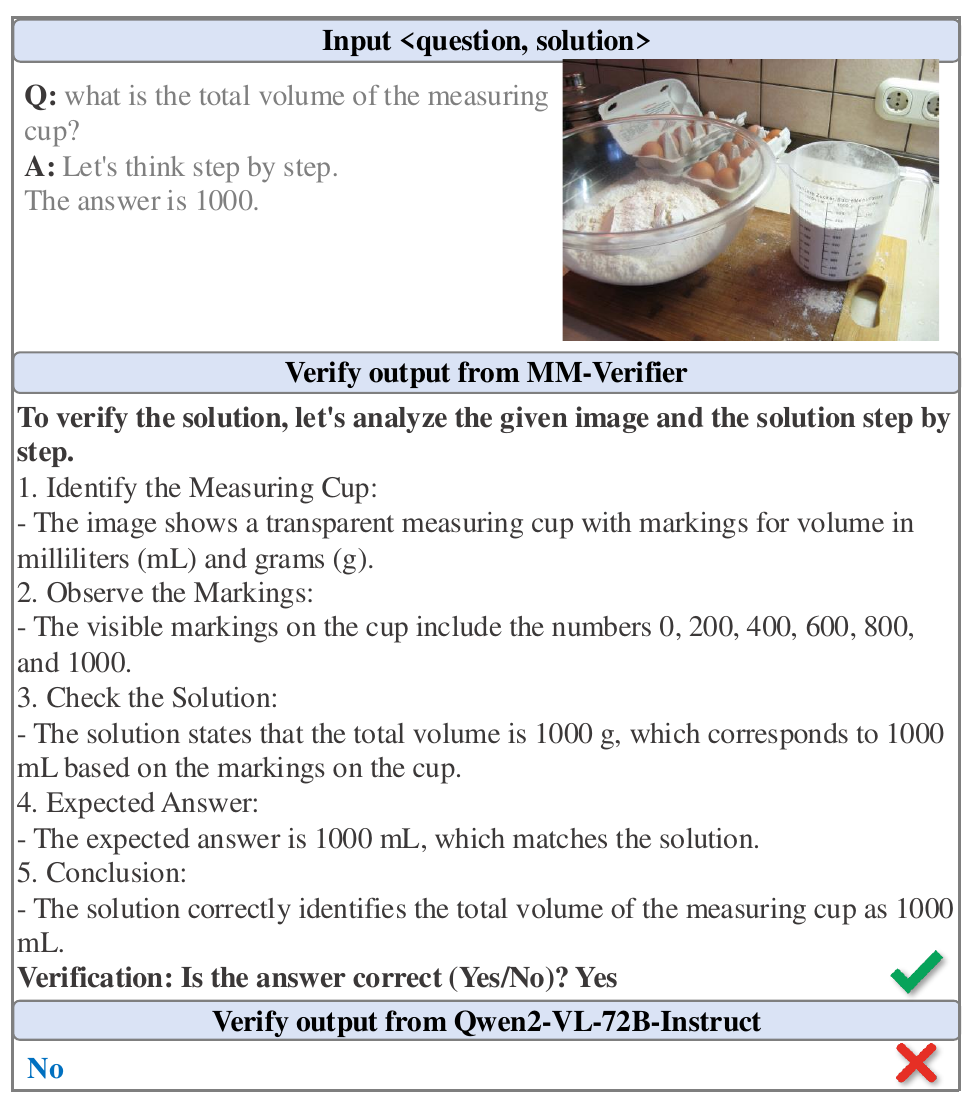}
\caption{We present case of MM-Verifier. We can see MM-Verifier correctly verify the answer with COT while Qwen2-VL-72B-Instruct failed to.}
\label{fig:case_appendix}
\vspace{-4mm}
\end{figure}

\section{Prompts}~\label{sec:Prompts}

\begin{figure*}[t]
\centering
\includegraphics[width=1\textwidth]{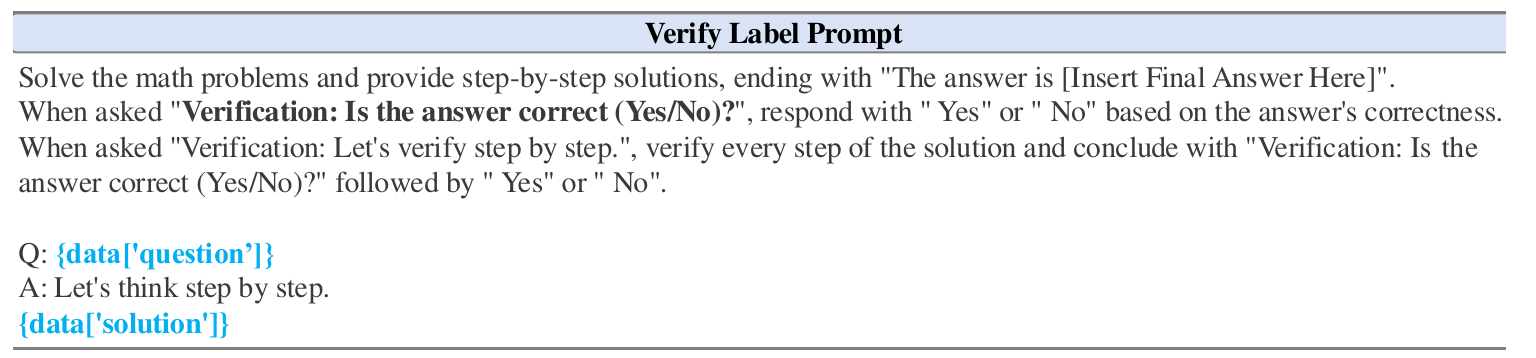}
\caption{Verify Label Prompt. This is the prompt we built with reference to~\citet{zhang2024generative}.}
\label{fig:prompt-verify}
\vspace{-2mm}
\end{figure*}

\begin{figure*}[t]
\centering
\includegraphics[width=1\textwidth]{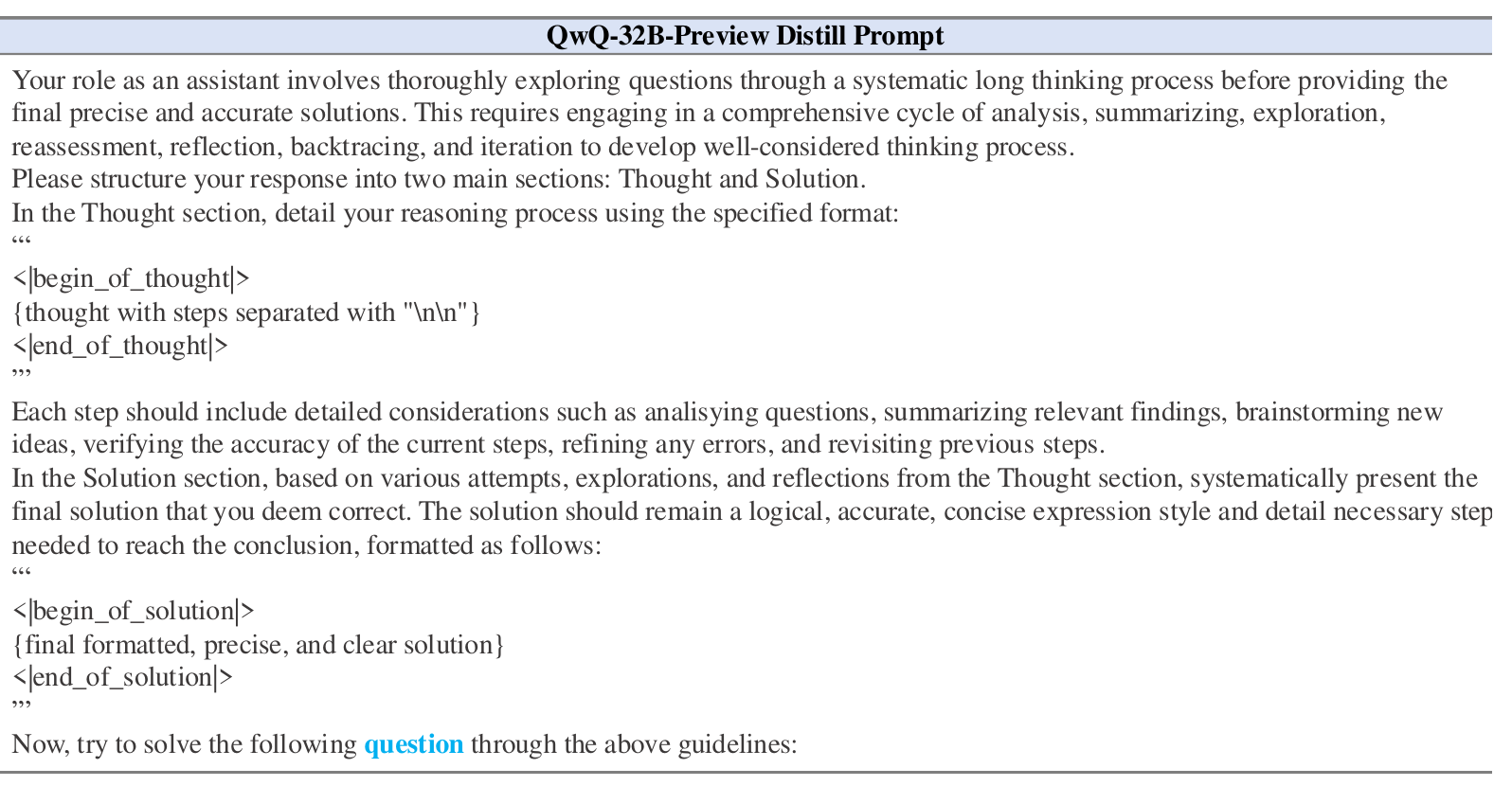}
\caption{QwQ-32B-Preview Distill Prompt. This is the prompt we built with reference to~\citet{min2024imitate}.}
\label{fig:prompt-qwq}
\vspace{-2mm}
\end{figure*}

\begin{figure*}[t]
\centering
\includegraphics[width=1\textwidth]{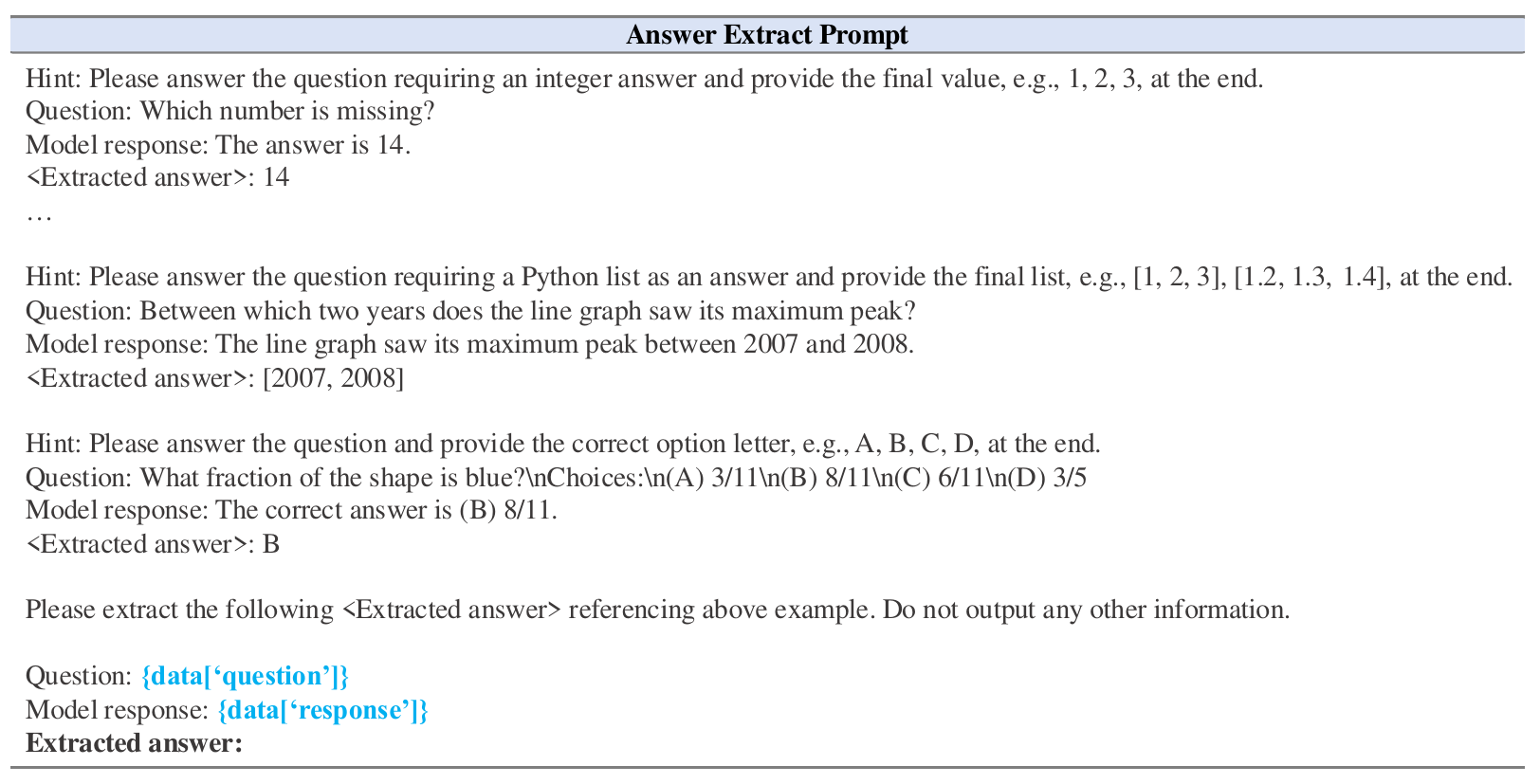}
\caption{Answer Extract Prompt}
\label{fig:prompt-extract}
\vspace{-2mm}
\end{figure*}

This paper primarily focuses on three key prompts: (1) the prompt for generating Verify data (Figure~\ref{fig:prompt-verify}), (2) the prompt (Figure~\ref{fig:prompt-qwq}) for distilling the QwQ-32B-Preview model after converting multimodal data into a textual format, and (3) the prompt for extracting answer data (Figure~\ref{fig:prompt-extract}).

\section{More Case Study}~\label{sec:appendix_case}
As shown in Figure~\ref{fig:case_appendix}, our MM-Verifier is capable of verifying the correctness of a solution that provides only a simple answer by leveraging CoT approach. In contrast, Qwen2-VL-72B-Instruct incorrectly classifies the solution as correct, highlighting its limitations in reasoning-based verification.

\end{document}